\documentclass[10pt,twocolumn,letterpaper]{article}

\makeatletter
\renewcommand\paragraph{\@startsection{paragraph}{4}{\z@}{1ex}{-1em}{\normalfont\normalsize\bfseries}}
\makeatother

\usepackage[dvipsnames,svgnames,x11names]{xcolor}
\usepackage{tikz}
\usetikzlibrary{arrows.meta,shapes,calc,matrix,fit,positioning,backgrounds,decorations.markings}
\usepackage{pgfplots}
\usepackage{pgfplotstable}
\pgfplotsset{compat=1.9}
\usepackage{xstring}

\usepgfplotslibrary{external}

\IfBeginWith*{\jobname}{fig/extern/}{\finalcopy}{}

\tikzstyle{every picture}+=[
	remember picture,
	every text node part/.style={align=center},
	every matrix/.append style={ampersand replacement=\&},
]
\tikzstyle{tight} = [inner sep=0pt,outer sep=0pt]
\tikzstyle{node}  = [draw,circle,tight,minimum size=12pt,anchor=center]
\tikzstyle{op}    = [draw,circle,tight]
\tikzstyle{dot}   = [fill,draw,circle,inner sep=1pt,outer sep=0]
\tikzstyle{pt}    = [fill,draw,circle,inner sep=1.5pt,outer sep=.2pt]
\tikzstyle{box}   = [draw,thick,rectangle,inner sep=3pt]
\tikzstyle{high}  = [black!60]
\tikzstyle{group} = [high,box,opacity=.5]
\tikzstyle{rectc} = [tight,transform shape]
\tikzstyle{rect}  = [rectc,anchor=south west]

\tikzset{every mark/.append style={solid}}
\pgfplotsset{
	grid=both, width=\columnwidth, try min ticks=5,
	every axis/.append style={font=\small},
	every axis plot/.append style={thick,mark=none,mark size=1.8,tension=0.18},
	legend cell align=left, legend style={fill opacity=0.8},
	xticklabel={\pgfmathprintnumber[assume math mode=true]{\tick}},
	yticklabel={\pgfmathprintnumber[assume math mode=true]{\tick}},
	nodes near coords math/.style={
		nodes near coords={\pgfmathprintnumber[assume math mode=true]{\pgfplotspointmeta}},
	},
}

\pgfplotsset{
	dash/.style={mark=o,dashed,opacity=0.6},
	dott/.style={mark=o,dotted,opacity=0.6},
	nolim/.style={enlargelimits=false},
	plain/.style={every axis plot/.append style={},nolim,grid=none},
}

\usepackage[accsupp]{axessibility}
\usepackage{wacv}
\usepackage{times}
\usepackage{epsfig}
\usepackage{graphicx}
\usepackage{amsmath}
\usepackage{amssymb}
\usepackage[utf8]{inputenc} 
\usepackage[T1]{fontenc}    
\usepackage{xspace}
\usepackage{bbm}            
\usepackage{url}
\usepackage{subfigure}
\usepackage{times}
\usepackage{adjustbox}
\usepackage{booktabs}
\usepackage{multirow}
\usepackage{enumitem}
\usepackage{xcolor}

\usepackage[ruled,vlined,linesnumbered]{algorithm2e}

\def\wacvPaperID{1179} 

\wacvfinalcopy 
\ifwacvfinal
\fi

\ifwacvfinal
\usepackage[breaklinks=true,bookmarks=false]{hyperref}
\else
\usepackage[pagebackref=true,breaklinks=true,colorlinks,bookmarks=false]{hyperref}

\fi

\ifwacvfinal
\usepackage[breaklinks=true,bookmarks=false]{hyperref}
\else
\usepackage[pagebackref=true,breaklinks=true,colorlinks,bookmarks=false]{hyperref}
\fi

\newcommand{\alert}[1]{{\color{red}{#1}}}

\newcommand{\eq}[1]{(\ref{eq:#1})}

\newcommand{\Th}[1]{\textsc{#1}}
\newcommand{\mr}[2]{\multirow{#1}{*}{#2}}
\newcommand{\mc}[2]{\multicolumn{#1}{c}{#2}}
\newcommand{\tb}[1]{\textbf{#1}}

\newcommand{\red}[1]{{\color{red}{#1}}}
\newcommand{\blue}[1]{{\color{blue}{#1}}}

\newcommand{\citeme}[1]{\red{[XX]}}
\newcommand{\refme}[1]{\red{(XX)}}

\newcommand{\fig}[2][1]{\includegraphics[width=#1\linewidth]{fig/#2}}

\newcommand{\tran}{^\top}

\newcommand{\real}{\mathbb{R}}

\newcommand{\normal}{\mathcal{N}}

\newcommand{\sigmoid}{\operatorname{sigmoid}}

\newcommand{\defn}{\mathrel{:=}}

\newcommand{\norm}[1]{\left\|{#1}\right\|}

\newcommand{\cN}{\mathcal{N}}

\newcommand{\cX}{\mathcal{X}}

\newcommand{\vzero}{\mathbf{0}}

\makeatletter
\newcommand*\bdot{\mathpalette\bdot@{.7}}
\newcommand*\bdot@[2]{\mathbin{\vcenter{\hbox{\scalebox{#2}{$\m@th#1\bullet$}}}}}
\makeatother

\makeatletter
\DeclareRobustCommand\onedot{\futurelet\@let@token\@onedot}
\def\@onedot{\ifx\@let@token.\else.\null\fi\xspace}

\def\eg{\emph{e.g}\onedot}

\makeatother

\newcommand{\base}{\mathrm{base}}
\newcommand{\novel}{\mathrm{novel}}
\newcommand{\CE}{\mathrm{CE}}
\newcommand{\KL}{\mathrm{KL}}
\newcommand{\KD}{\mathrm{KD}}
\newcommand{\hal}{\mathrm{H}}

\begin{document}

\title{Tensor feature hallucination for few-shot learning}

\author{
Michalis Lazarou$^1$ \ \ \ \ Tania Stathaki$^1$ \ \ \ \ Yannis Avrithis$^2$\\
$^1$Imperial College London\\
$^2$Athena RC\\
}

\maketitle

\begin{abstract}
Few-shot learning addresses the challenge of learning how to address novel tasks given not just limited supervision but limited data as well.  An attractive solution is synthetic data generation. However, most such methods are overly sophisticated, focusing on high-quality, realistic data in the input space. It is unclear whether adapting them to the few-shot regime and using them for the downstream task of classification is the right approach. Previous works on synthetic data generation for few-shot classification focus on exploiting complex models, \eg a Wasserstein GAN with multiple regularizers or a network that transfers latent diversities from known to novel classes.

We follow a different approach and investigate how a simple and straightforward synthetic data generation method can be used effectively. We make two contributions, namely we show that: (1) using a simple loss function is more than enough for training a feature generator in the few-shot setting; and (2) learning to generate tensor features instead of vector features is superior. Extensive experiments on \emph{mini}Imagenet, CUB and CIFAR-FS datasets show that our method sets a new state of the art, outperforming more sophisticated few-shot data augmentation methods. The source code can be found at \url{https://github.com/MichalisLazarou/TFH_fewshot}.

\end{abstract}

\section{Introduction}
\label{intro}

Deep learning continuously improves the state of the art in different fields, such as \emph{natural language understanding}~\cite{language} and \emph{computer vision}~\cite{imagenet}. However, a fundamental limitation of representation learning from raw data is the dependence on large amounts of task-specific or domain-specific data, labeled or not. This limitation inhibits the application of deep learning to real-world problems, such as rare species classification, where the cost of obtaining and annotating data from a new domain is high.

\begin{figure}[h]
\centering
\fig{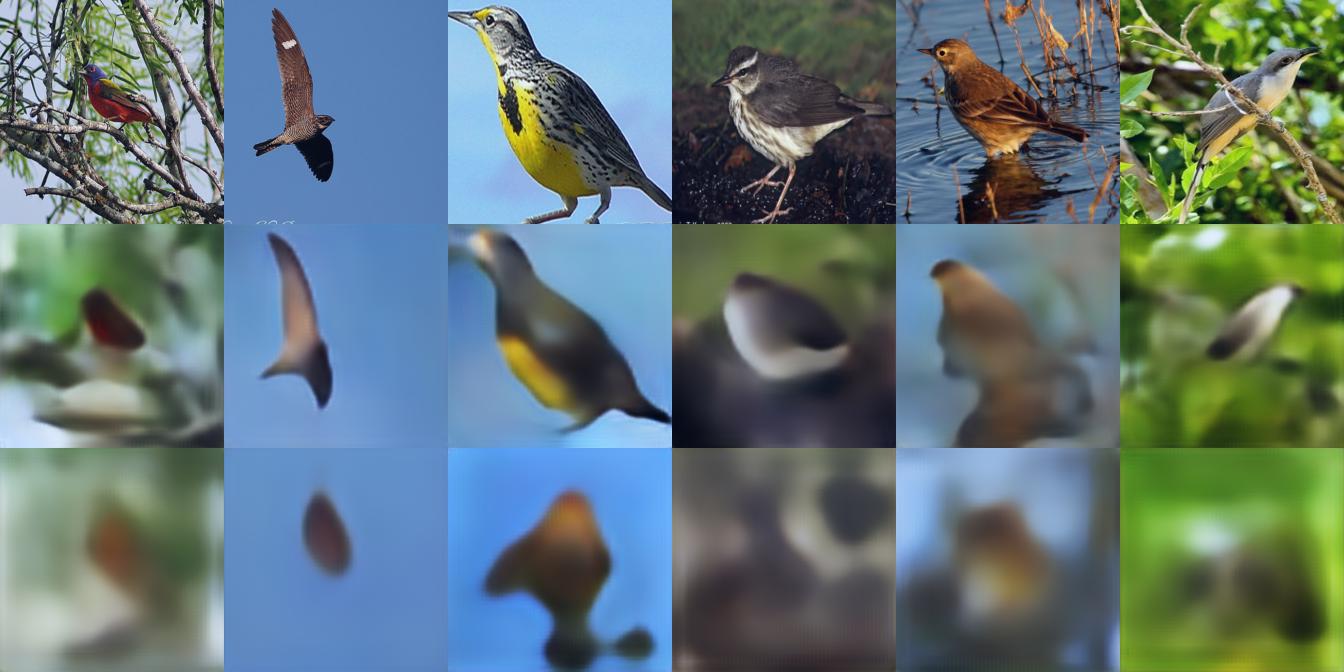}
\caption{CUB original images (row 1) followed by images generated from separately trained reconstructors using as input tensor features (row 2) or vector features (row 3). More results and implementation details are given in the supplementary material.}
\label{fig:im_recon}
\end{figure}

To address this limitation, \emph{few-shot learning}~\cite{matchingNets, prototypical, MAML} has attracted significant interest in recent years. Few-shot learning is concerned with learning not only under limited supervision, but also from limited data. This constraint excludes representation learning from scratch and inhibits adapting the representation, which is otherwise common in \emph{transfer learning}~\cite{donahue2014decaf,ECCV2020_211} \emph{domain/task adaptation}~\cite{ganin2015unsupervised, rebuffi2017learning} and \emph{continual learning}~\cite{rebuffi2017icarl}.

\begin{figure*}[ht]
\centering
\fig{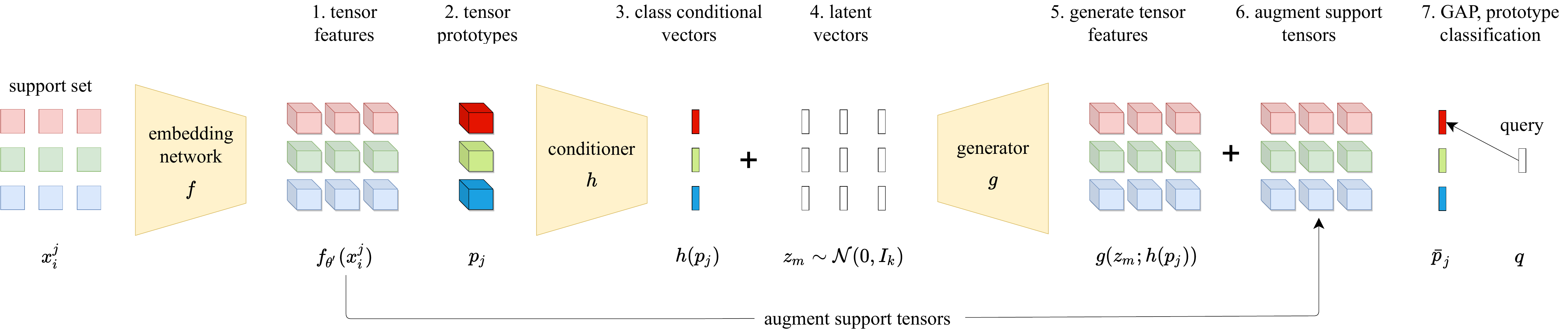}
\vspace{0pt}
\caption{\emph{Overview of our method}. \emph{At inference}: 1) Map the support examples $x_i^j$ (each color indicates a different class $j$) into tensor features $f_{\theta'}(x_i^j)$ through the pre-trained embedding network $f_{\theta'}$. 2) Average $x_i^j$ into a tensor prototype $p_j$ per class $j$~\eq{proto}. 3) Map each $p_j$ to a class conditional vector $h(p_j)$ through the \emph{conditioner network} $h$. Draw $M$ samples $z_m$ per class from a $k$-dimensional normal distribution $\cN(\vzero, I_k)$. 5) Generate $M$ class-conditional tensor features $g(z_m; h(p_j))$ per class $j$ using \emph{generator network} $g$. 6) Augment the support tensors with the generated tensors. 7) Perform \emph{global average pooling} (GAP) and average the augmented features into vector prototype $\bar{p}_j$ per class $j$~\eq{proto2} and classify queries $q$ to nearest prototype. \emph{At training} (not shown): a) Train $f_\theta$ using cross-entropy~\eq{cls}. b) Fine-tune $f_\theta$ to $f_{\theta'}$ using self-distillation~\eq{kl}. c) Train \emph{tensor feature hallucinator} (TFH) $\{h, g\}$ using reconstruction loss~\eq{mse}.}
\label{fig:overview}
\end{figure*}

\emph{Data augmentation}, commonly based on simple input transformations, is a universal way of regularizing and improving the generalization ability of a model~\cite{imagenet}, as well as exploiting unlabeled data~\cite{chen2020simple, sohn2020fixmatch}. In few-shot learning, recent methods go beyond input transformations towards \emph{synthetic data generation} and \emph{hallucination}, either in the \emph{image space}~\cite{Ideme, metaGAN} or in the \emph{feature space}~\cite{semanticAugmentation, AFHN, VIFSL}. Hence, they address the data deficiency by augmenting real data with synthetic, achieving a greater extent of diversity.

The vast majority of generative models focuses on high-quality, high-resolution images, assuming a large amount of data. Also, the metrics used to evaluate generative models, focus on whether the generated data is realistic~\cite{FID, IS}. Generating high quality, realistic data may not be necessary in ``downstream tasks'' such as classification. It is unclear whether and how state of the art generative models in the image space can succeed in the few-shot setting.

Most of the recent few-shot feature hallucination methods focus on generating vectors in the feature space~\cite{semanticAugmentation, AFHN, VIFSL}. These vectors are most commonly obtained by \emph{global average pooling} (GAP) on the output feature maps. This discards spatial details that might be necessary to model the underlying data distribution. We hypothesize that working with the feature map tensors directly may be more effective. To investigate this, we train two image reconstructors separately: one for tensor features and the other for vector features obtained by GAP. The latter has the same architecture as the former, except for one additional upsampling layer. As shown in \autoref{fig:im_recon}, the feature map tensors preserve more information indeed.

Motivated by this finding, we explore the potential of using \emph{tensor features} instead of vector features in a simple generative model to improve few-shot classification. We employ a simple \emph{conditioner}-\emph{generator} architecture and we introduce a simple \emph{reconstruction loss} between the generated tensor and the corresponding class prototype tensor. This allows the generation of a diverse set of synthetic data, not necessarily realistic, from a limited amount of real data from a previously unseen task. An overview is shown in \autoref{fig:overview}. We demonstrate empirically that our model provides state of the art results, outperforming more sophisticated generative models on a number of benchmarks. Our contributions are summarized as follows:

\begin{enumerate}[itemsep=3pt, parsep=0pt, topsep=0pt]
    \item We are the first to generate \emph{tensor features} instead of vector features in the few-shot setting and to leverage their structural properties (\autoref{sec:tensor_hallucination}).

    \item We introduce a novel loss function that is simpler than alternatives in state of the art few-shot synthetic data generation methods~\cite{AFHN, VIFSL, DTN} (\autoref{sec:training_loss}).

    \item Our \emph{tensor feature hallucinator} (TFH) sets new state of the art on three common few-shot classification benchmarks: \emph{mini}Imagenet, CUB and CIFAR-FS.

    \item We demonstrate the robustness of our hallucinator against using different backbone networks and classifiers, as well as its applicability to the challenging setting of \emph{cross-domain} few-shot learning.
\end{enumerate}

\section{Related work}
\label{sec:related}

\subsection{Few-shot learning}
\label{sec:fsl}

In few-shot learning, the objective is to learn from an abundant labeled set of \emph{base} classes how to solve tasks from a limited \emph{support} set over a distinct set of \emph{novel} classes. We briefly discuss different approaches to this objective, followed by a more detailed account of \emph{synthetic data generation}, where our contribution lies.

\paragraph{Meta-learning}

The objective of few-shot classification fits naturally within \emph{meta-learning}~\cite{schmidhuber1987evolutionary,vilalta2002perspective}, referring to learning at two levels, where generic knowledge is acquired before adapting to more specific tasks. There are different instantiations of this idea, all sharing the fact that the loss is computed on a \emph{set} of examples, called an \emph{episode}.

\emph{Optimization meta-learning} aims to learn how to quickly update the model parameters to novel tasks without overfitting. This includes updates in closed form~\cite{bertinetto}, iterative updates according to the gradient descent~\cite{MAML, CAVIA, reptile} and learnable iterative updates, such as LSTM~\cite{ravilstm}.

\emph{Model-based meta-learning} aims to learn how to update specific model architectures to novel tasks. This includes memory-augmented neural networks~\cite{santoro} and meta-networks, designed explicitly according to the two-level learning paradigm~\cite{metaNet}.

\emph{Metric learning} is a standalone field that overlaps meta-learning and aims to learn how to compare examples of unseen classes~\cite{oh2016deep}. Most often, it amounts to learning an \emph{embedding space} where distances or similarities are taken between a query and individual examples~\cite{siamese}, all examples of a class~\cite{matchingNets}, or the class centroid~\cite{prototypical}. A metric or similarity function may also be learned directly~\cite{relationnet}.

\paragraph{Representation learning}

Instead of learning in episodes, it is simpler to compute the loss on one example at a time, like standard cross-entropy. Learning a classifier on the base classes then amounts to \emph{representation learning}. This simplified approach has been popularized in few-shot classification with the \emph{cosine-based classifier}~\cite{imprintedweights, gidarisFSL, denseclassification, closerlook}. Any method that helps in learning a better representation is applicable in this sense, including pretext tasks~\cite{gidarisRot}, self-distillation~\cite{rfs} and manifold mixup~\cite{manifoldmixup}.

\paragraph{Task adaptation}

In few-shot learning, the challenge is to adapt the representation to novel tasks on limited data without overfitting. This is possible using metric scaling~\cite{tadam}, attention mechanisms~\cite{gidarisFSL}, task-adaptive projections~\cite{tapnet}, set-to-set functions~\cite{feat}, identifying task-relevant features~\cite{categorytraversalFSL} or growing the architecture~\cite{denseclassification}.


\paragraph{Unlabeled data}

The constraint of few-shot learning may be relaxed by accessing more novel-class unlabeled data (\emph{semi-supervised learning}) or multiple queries at a time (\emph{transductive inference}). This allows exploiting the \emph{manifold structure} of the additional data, for instance using graph neural networks~\cite{fewshotGNN1, fewshotGNN2}, label propagation~\cite{tpn}, embedding propagation~\cite{embeddingpropagation} or label cleaning~\cite{ilpc}.


\subsection{Synthetic data generation}
\label{sec:synth}

Generative models aim to model the underlying data distribution of the training set in a \emph{latent space}. \emph{Generative adversarial networks} (GAN)~\cite{GANSgoodfellow} are by far the most popular approach. The idea is a zero-sum game between a generator and a discriminator, such that generated data are realistic enough to be indistinguishable from real. Several improvements concern the architecture, as well as training and regularization methods \cite{stylegan, biggan, pggan}.

Other approaches include \emph{variational autoencoders} (VAE)~\cite{VAE}, imposing a prior distribution in the latent space, \emph{autoregressive} (AR) \emph{models} \cite{pixelCNN, condpixelCNN}, iteratively generating samples conditioned on previous steps, and \emph{flow-based models} \cite{glow, flow++}, modeling the data distribution explicitly via invertible transformations.

Most state of the art generative models do not focus on improving the performance of downstream tasks such as classification, but rather on image quality metrics such as \emph{Fréchet inception distance} (FID)~\cite{FID} and \emph{inception score} (IS)~\cite{IS}. They also assume access to abundant training data, which is in direct contrast to the few-shot setting. Even though recent methods address small datasets \cite{styleganADA, few-shotGAN}, they are limited to unconditional image generation, which inhibits their use in novel tasks.


\subsection{Synthetic data generation for few-shot learning}
\label{sec:synth-fsl}

In few-shot learning, a generative model can be learned on base-class data to augment real novel-class data with synthetic. One of the first ideas in this direction is \emph{feature hallucination}~\cite{lowshot2}. There are several approaches based on GANs, including MetaGAN~\cite{metaGAN}, which integrates MAML~\cite{MAML} with a conditional GAN to generate examples in the input space; AFHN~\cite{AFHN}, a feature hallucinator using wGAN~\cite{wGAN}; and FUNIT~\cite{funit}, a GAN-based method for few-shot image-to-image translation.


There are also alternative approaches to GANs, including VI-Net~\cite{VIFSL}, which uses a class-conditional VAE as a feature hallucinator; \emph{diversity transfer network} (DTN)~\cite{DTN}, which learns how to transfer latent diversities from base to novel classes; SalNet~\cite{salnetFSL}, which hallucinates features by combining foregrounds and backgrounds from different images; and IDeMe-Net~\cite{Ideme}, which combines novel-class support with similar base-class images.

All the aforementioned methods, including the current state of the art, use overly complex training regimes, adapting them to the few-shot setting. For instance, AFHN~\cite{AFHN} adapts the Wasserstein GAN~\cite{wGAN} and VI-Net~\cite{VIFSL} adapts a variational autoencoder~\cite{VAE}. It is not clear whether such sophisticated generative models and loss functions are necessary for a ``downstream task'' like few-shot classification. At the same time, generating vector features incurs information loss as demonstrated in \autoref{fig:im_recon}.

\subsection{On our contribution}
\label{sec:contr}

Our work falls within generating synthetic data in the \emph{feature space}. However, we are the first to train a model to generate \emph{tensor features} instead of vector features in few-shot learning, exploiting the spatial and structural properties of tensors (\autoref{sec:tensor_hallucination}). This allows us to use a \emph{simple reconstruction loss} between the generated tensors and their class prototype (\autoref{sec:training_loss}), while still outperforming methods using overly complex generative models.

Our model bears similarities to a VAE~\cite{VAE}, also used by VI-Net~\cite{VIFSL}. Our \emph{conditioner} and \emph{hallucinator} networks play a similar role to encoder and decoder, respectively. However, rather than predicting the variance and imposing a prior distribution in the latent space, we condition the model on \emph{class prototypes}, also represented by tensor features, and we use the same prototypes in the reconstruction loss.

To improve representation learning, we use \emph{self-distillation} as an auxiliary loss term, following~\cite{rfs} (\autoref{sec:representation}). We also perform task adaptation by \emph{fine-tuning} the hallucinator to novel class data for few iterations.


\section{Method}
\label{sec:method}

\subsection{Problem formulation}
\label{sec:problem}

We are given a labeled dataset $D_{\base} \defn \{(x_i, y_i)\}_{i=1}^I$ of $I$ examples, with each example $x_i$ having a label $y_i$ in one of the classes in $C_{\base}$. This dataset is used to learn the parameters $\theta$ of a mapping $f_\theta: \cX \to \real^{d \times h \times w}$ from an input image space $\cX$ to a \emph{feature} or \emph{embedding} space, where \emph{feature tensors} have $d$ dimensions (channels) and spatial resolution $h \times w$ (height $\times$ width).

The knowledge acquired at representation learning is used to solve \emph{novel tasks}, assuming access to a dataset $D_{\novel}$, with each example being associated with one of the classes in $C_{\novel}$, where $C_{\novel}$ is disjoint from $C_{\base}$. In \emph{few-shot classification}~\cite{matchingNets}, a novel task is defined by sampling a \emph{support set} $S$ from $D_{\novel}$, consisting of $N$ classes with $K$ labeled examples per class, for a total of $L \defn NK$ examples. Given the mapping $f_\theta$ and the support set $S$, the problem is to learn an $N$-way classifier that makes predictions on unlabeled \emph{queries}, also sampled from novel classes. Queries are treated independently of each other. This is referred to as \emph{inductive inference}.


\subsection{Representation learning}
\label{sec:representation}

The goal of \emph{representation learning} is to learn the embedding function $f_\theta$ that can be applied to $D_{\novel}$ to extract embeddings and solve novel tasks. We use $f_\theta$ followed by \emph{global average pooling} (GAP) and a parametric \emph{base classifier} $c_\phi$ to learn the representation. We denote by $\bar{f}_\theta: \cX \to \real^d$ the composition of $f_\theta$ and GAP. We follow the two-stage regime by~\cite{rfs} to train our embedding model. In the \emph{first stage}, we train $f_\theta$ on $D_{\base}$ using standard cross-entropy loss $L_{\CE}$:
\begin{equation}
    J(D_{\base}; \theta, \phi) \defn
		\sum_{i=1}^I \ell_{\CE}(c_{\phi}(\bar{f}_\theta(x_i)), y_i) + R(\phi),
\label{eq:cls}
\end{equation}
where $R$ is a regularization term. In the \emph{second stage}, we adopt a \emph{self-distillation} process: The embedding model $f_\theta$ and classifier $c_\phi$ from the first stage serve as the teacher and we distill their knowledge to a new student model $f_{\theta'}$ and classifier $c_{\phi'}$, with identical architecture. The student is trained using a linear combination of the standard cross-entropy loss, as in stage one, and the Kullback-Leibler (KL) divergence between the student and teacher predictions:
\begin{equation}
\begin{split}
    J_{\KD} & (D_{\base}; \theta', \phi') \defn \\
		& \alpha J(D_{\base}; \theta', \phi') + \\
		& \beta \KL(c_{\phi'}(\bar{f}_{\theta'}(x_i)), c_\phi(\bar{f}_{\theta}(x_i))),
\label{eq:kl}
\end{split}
\end{equation}
where $\alpha$ and $\beta$ are scalar weights and $\theta, \phi$ are fixed.


\subsection{Feature tensor hallucinator}
\label{sec:tensor_hallucination}

Existing feature hallucination methods \cite{AFHN, DTN, VIFSL, salnet, lowshot2} are trained using \emph{vector features}, losing significant spatial and structural information. By contrast, our hallucinator is trained on tensor features before GAP and generates \emph{tensor features} as well. In particular, we use the student model $f_{\theta'}: \cX \to \real^{d \times h \times w}$, pre-trained using \eq{kl}, as our embedding network to train our tensor feature hallucinator.

The hallucinator consists of two networks: a \emph{conditioner} network $h$ and a \emph{generator} network $g$. The conditioner aids the generator in generating class-conditional examples. Given a set $X_j \defn \{x_i^j\}_{i=1}^K$ of examples associated with each class $j=1,\dots,N$, conditioning is based on the \emph{prototype tensor} $p_j \defn p(X_j) \in \real^{d \times h \times w}$ of class $j$,
\begin{equation}
    p(X_j) \defn \frac{1}{K} \sum_{i=1}^{K} f_{\theta'}(x_i^j).
\label{eq:proto}
\end{equation}
The conditioner $h: \real^{d \times h \times w} \to \real^{d'}$ maps the prototype tensor to the \emph{class-conditional vector} $s_j \defn h(p_j) \in \real^{d'}$. The generator $g: \real^{k+d'} \to \real^{d \times h \times w}$ takes as input this vector as well as a \emph{latent vector} $z \sim \normal(\vzero,I_k)$ drawn from a $k$-dimensional standard normal distribution and generates a \emph{class-conditional tensor feature} $g(z; s_j) \in \real^{d \times h \times w}$ for each class $j$.

\begin{algorithm}
\footnotesize
\DontPrintSemicolon
\SetFuncSty{textsc}
\SetDataSty{emph}
\newcommand{\commentsty}[1]{\blue{#1}}
\SetCommentSty{commentsty}
\SetKwComment{Comment}{$\triangleright$ }{}

\SetKwInOut{Input}{input}
\SetKwInOut{Output}{output}

\Input{training set $D_{\base}$}
\Input{pre-trained embedding $f_{\theta}$}
\Output{trained tensor hallucinator $\{h, g\}$}
\BlankLine

\While{not done}
{
	Sample an $N$-way $K$-shot episode $E \defn \{E_j\}_{j=1}^N$ from $D_{\base}$ \\
	\For{class $j=1,\dots,N$}
	{
		Obtain class prototype tensor $p_j \defn p(E_j)$ by~\eq{proto} \\
		Map $p_j$ to class-conditional vector $s_j \defn h(p_j)$ \\
		Draw $M$ samples $\{z_m\}^M_{m=1}$ from $\normal(\vzero,I_k)$ \\
		Generate $M$ class-conditional tensor features $\{g(z_m; s_j)\}_{m=1}^M$ \\
	}
	Update parameters of hallucinator $\{h, g\}$ by~\eq{mse}
}
\caption{Meta-training of tensor hallucinator}
\label{alg:hal_training}
\end{algorithm}


\subsection{Training the hallucinator}
\label{sec:training_loss}

We train our hallucinator using a meta-training regime, similar to \cite{AFHN, semanticAugmentation, deltaencoder}. At every iteration, we sample a new episode by randomly sampling $N$ classes and $K$ examples $E_j \defn \{x_i^j\}_{i=1}^K$ for each class $j$ from $D_{\base}$. For each class $j=1,\dots,N$, we obtain the prototype tensor $p_j \defn p(E_j)$ using~\eq{proto} and the class-conditional vector $s_j \defn h(p_j)$ by the conditioner $h$. We then draw $M$ samples $\{z_m\}_{m=1}^M$ from the standard normal distribution $\normal(\vzero,I_k)$ and generate $M$ class-conditional tensor features $\{g(z_m; s_j)\}_{m=1}^M$ using the generator $g$. We train our hallucinator $\{h,g\}$ on the episode data $E \defn \{E_j\}_{j=1}^N$ by minimizing the \emph{mean squared error} (MSE) of generated class-conditional tensor features of class $j$ to the corresponding class prototype $p_j$:
\begin{equation}
	J_{\hal}(E; h, g) = \frac{1}{MN} \sum_{j=1}^N \sum_{m=1}^M \norm{g(z_m; h(p_j)) - p_j}^2.
\label{eq:mse}
\end{equation}
Algorithm \ref{alg:hal_training} summarizes the overall training process of the hallucinator.

\begin{algorithm}
\footnotesize

\DontPrintSemicolon
\SetFuncSty{textsc}
\SetDataSty{emph}
\newcommand{\commentsty}[1]{\blue{#1}}
\SetCommentSty{commentsty}
\SetKwComment{Comment}{$\triangleright$ }{}

\SetKwInOut{Input}{input}
\SetKwInOut{Output}{output}

\Input{support set $S \defn \{S_j\}_{j=1}^N$}
\Input{pre-trained embedding $f_{\theta}$}
\Input{pre-trained hallucinator $\{h, g\}$}
\Output{predicted label for query $q \in \cX$}
\BlankLine

\For{class $j=1,\dots,N$}
{
	Obtain class prototype tensor $p_j \defn p(S_j)$ by~\eq{proto} \\
	Draw $M$ samples $\{z_m\}^M_{m=1}$ from $\normal(\vzero,I_k)$ \\
	Generate $M$ class-conditional tensor features $G_j \defn \{g(z_m; h(p_j))\}_{m=1}^M$ \\
	Augment the support $S_j$ with generated features $G_j$ \\
	Apply GAP and obtain vector class prototypes $\bar{p}_j$ by~\eq{proto2}
}
Assign query $q$ to class of nearest prototype to feature $\bar{f}_{\theta'}(q)$

\caption{Using hallucinator at inference}
\label{alg:hal_inference}
\end{algorithm}


\subsection{Inference}

At inference, we are given a few-shot task with a support set $S \defn \{S_j\}_{j=1}^N$, containing $N$ novel classes with $K$ examples $S_j \defn \{x_i^j\}_{i=1}^K$ for each class $j$. For each class $j=1,\dots,N$, we use our trained backbone network $f_{\theta'}$ to compute the tensor feature $f_{\theta'}(x_i^j) \in \real^{d \times h \times w}$ of each example in $S_j$ and we obtain the prototype $p_j \defn p(S_j)$ by~\eq{proto}. Then, using our trained tensor feature hallucinator $\{h,g\}$, we generate $M$ class-conditional tensor features $G_j \defn \{g(z_m; h(p_j))\}_{m=1}^M$, also in $\real^{d \times h \times w}$, where $z_m$ are drawn from $\normal(\vzero,I_k)$. We augment the support features $f_{\theta'}(S_j)$ with the generated features $G_j$, resulting in $K+M$ labeled tensor features per class in total. We now apply GAP to those tensor features and obtain new, \emph{vector class prototypes} in $\real^d$:
\begin{equation}
    \bar{p}_j \defn \frac{1}{K+M} \left(
			\sum_{i=1}^{K} \bar{f}_{\theta'}(x_i^j) +
			\sum_{m=1}^M \bar{g}(z_m; h(p_j))
		\right),
\label{eq:proto2}
\end{equation}
where $\bar{g}$ denotes the composition of $g$ and GAP. Finally, given a query $q \in \cX$, we apply GAP to the tensor feature $f_{\theta'}(q)$ and assign it to the class of the nearest vector prototype. Algorithm \ref{alg:hal_inference} summarizes the inference process.

We refer to the above approach as \emph{prototypical classifier}. In \autoref{sec:ablation} we experiment with alternative classifiers such as logistic regression and support vector machine on the same augmented (support $+$ generated) features.

\section{Experiments}
\label{experiments}

\newcommand{\ci}[1]{{\tiny $\pm$#1}}
\newcommand{\cip}{\phantom{\ci{0.00}}}
\newcommand{\cim}{\ci{\alert{0.00}}}

\subsection{Datasets}
We use
three common few-shot classification datasets: \emph{mini}ImageNet~\cite{matchingNets, ravilstm}, CUB~\cite{closerlook, fewshotCUB} and CIFAR-FS~\cite{closerlook, cifar100db}. More details are given in the supplementary material.
\begin{figure}
\small
\centering
\setlength\tabcolsep{4pt}
\input{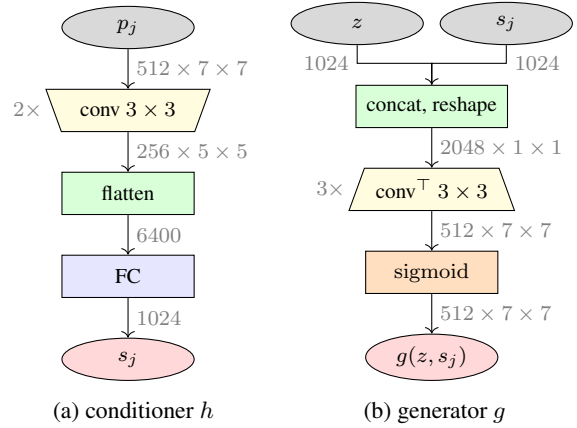}
\begin{tabular}{cc}
\begin{tikzpicture}[
	font={\footnotesize},
]
\matrix[
	row sep=15pt,column sep=10pt,cells={scale=1,},
]{
	\node[inp](p){$p_j$}; \\
	\node[down](d){conv $3 \times 3$}; \\
	\node[manip](f){flatten}; \\
	\node[fix](fc){FC}; \\
	\node[outp](s){$s_j$}; \\
};
\draw[->]
	(p) edge node[dim,midway,right]{$512 \times 7 \times 7$} (d)
	(d) edge node[dim,midway,right]{$256 \times 5 \times 5$} (f)
	(f) edge node[dim,midway,right]{$6400$} (fc)
	(fc) edge node[dim,midway,right]{$1024$} (s)
	;
\node[dim,left=1pt of d] {$2\times$};
\end{tikzpicture} &
\begin{tikzpicture}[
	font={\footnotesize},
]
\matrix[
	row sep=15pt,column sep=10pt,cells={scale=1,},
]{
	\coordinate(O) at(0,0);
	\node[inp,left=3pt](z){$z$};
	\node[inp,right=3pt](s){$s_j$}; \\
	\node[manip](c){concat, reshape}; \\
	\node[up](ct){conv$\tran$ $3 \times 3$}; \\
	\node[fun](S){$\sigmoid$}; \\
	\node[outp](o){$g(z, s_j)$}; \\
};
\coordinate(m) at($(O)!.65!(c.north)$); 
\draw[->]
	(c) edge node[dim,midway,right]{$2048 \times 1 \times 1$} (ct)
	(ct) edge node[dim,midway,right]{$512 \times 7 \times 7$} (S)
	(S) edge node[dim,midway,right]{$512 \times 7 \times 7$} (o)
	;
\node[dim,left=1pt of ct] {$3\times$};
\draw[->]
	(z) |- node[dim,midway,left]{$1024$} (m) -- (c)
	(s) |- node[dim,midway,right]{$1024$} (m) -- (c);
\end{tikzpicture} \\
(a) conditioner $h$ &
(b) generator $g$
\end{tabular}
\vspace{6pt}
\caption{Architecture of our (a) conditioner network $h$, (b) generator network $g$ for ResNet-18. See text for more details and for ResNet-12. conv$\tran$: transpose convolution.}
\label{fig:arch}
\end{figure}


\subsection{Networks}

Our \emph{tensor feature hallucinator} (TFH) consists of a conditioner network and a generator network. Their architecture depends on the backbone network.

\paragraph{Backbone}

Many recent data augmentation methods \cite{semanticAugmentation, Ideme, AFHN, VIFSL} use ResNet-18 \cite{resnet18} as a backbone embedding model. To perform as fair comparison as possible, we use this backbone by default. To investigate the transferability and robustness of our tensor hallucinator, we also use a pre-trained ResNet-12 backbone from the publicly available code of DeepEMD~\cite{deepEMD}.

For ResNet-18, the embedding dimension is $d = 512$ and the resolution $h \times w = 7 \times 7$. For ResNet-12, it is $d = 640$ and $h \times w = 5 \times 5$.

\paragraph{Conditioner}

As shown in \autoref{fig:arch}(a), our conditioner $h: \real^{d \times h \times w} \to \real^{d'}$ consists of two convolutional layers with a ReLU activation in-between, followed by flattening and a fully-connected layer. The convolutional layers use kernels of size $3 \times 3$, and stride 1. In the first convolutional layer, we also use padding 1. The output channels are $d$ and $d/2$ in the first and second layer, respectively. The dimension of the \emph{class-conditional vector} is $d' = 1024$.

For ResNet-18, the tensor dimensions of all conditioner layers are $[512 \times 7 \times 7]$, $[512 \times 7 \times 7]$, $[256 \times 5 \times 5]$, $[6400]$ (flattening) and $[1024]$. For ResNet-12, they are $[640 \times 5 \times 5]$, $[640 \times 5 \times 5]$, $[320 \times 3 \times 3]$, $[2880]$ (flattening) and $[1024]$.

\paragraph{Generator}

As shown in \autoref{fig:arch}(b), for ResNet-18, our generator $g: \real^{k+d'} \to \real^{d \times h \times w}$ consists of concatenation of $z$ and $s_j$ into $(z; s_j) \in \real^{k+d'}$, followed by reshaping to $(k+d') \times 1 \times 1$, three transpose-convolutional layers with ReLU activations in-between and a sigmoid function at the end. For ResNet-12, the generator architecture is the same, except that it has only \emph{two} transpose-convolutional layers. The dimension of the latent vector $z$ is $k = 1024$. All transpose-convolutional layers use kernels of size $3 \times 3$, stride 1 and $d$ output channels.

For ResNet-18, the tensor dimensions of all generator layers are $[2048 \times 1 \times 1]$, $[512 \times 3 \times 3]$, $[512 \times 5 \times 5]$, and $[512 \times 7 \times 7]$. For ResNet-12, they are $[2048 \times 1 \times 1]$, $[640 \times 3 \times 3]$, and $[640 \times 5 \times 5]$.




\subsection{Training}
\label{sec:train}

\paragraph{Embedding model}

Similarly to \cite{rfs}, we use SGD optimizer with learning rate 0.05, momentum 0.9 and weight decay 0.0005. For data augmentation, as in \cite{metaridge}, we adopt random crop, color jittering, and horizontal flip.

\paragraph{Tensor feature hallucinator (TFH)}

Our TFH is trained in episodes of $N = 5$ classes, $K = 20$ examples per class and  generation of $M = 50$ class-conditional examples. We train for 50 epochs, where each epoch consists of 600 episodes. We use Adam optimizer with initial learning rate $10^{-4}$, decaying by half at every 10 epochs.

\paragraph{Novel-task fine-tuning (TFH-ft)}

Given a novel task, we also provide an improved solution, TFH-ft, where our hallucinator is \emph{fine-tuned} on novel-class support examples. We use exactly the same loss function as in hallucinator training~\eq{mse} and we fine-tune for $t$ steps using Adam optimizer and learning rate $\eta$. \autoref{tab:ft} shows the values $t$ and $\eta$ used in all experiments.

\begin{table}
\small
\centering
\setlength\tabcolsep{4pt}
\begin{tabular}{lcccccc} \toprule
\mr{2}{\Th{Backbone}} & \mc{2}{\Th{1-shot}} & \mc{2}{\Th{5-shot}} \\ \cmidrule{2-5}
                      & $k$ & $\gamma$      & $k$ & $\gamma$      \\ \midrule
ResNet-18             & 15  & $10^{-7}$     & 10  & $10^{-4}$     \\
ResNet-12             & 10   & $10^{-7}$     & 10   & $10^{-4}$     \\ \bottomrule
\end{tabular}
\vspace{6pt}
\caption{Number of steps $t$ and learning rate $\eta$ for TFH-ft, chosen on \emph{mini}ImageNet validation set and used in all experiments.}
\label{tab:ft}
\end{table}

\begin{table*}
\small
\centering
\setlength\tabcolsep{4pt}
\begin{tabular}{lccccccc} \toprule
\mr{2}{\Th{Method}}                            & \mr{2}{\Th{Backbone}} & \mc{2}{\Th{\emph{mini}ImageNet}}          & \mc{2}{\Th{CUB}}                          & \mc{2}{\Th{CIFAR-FS}}                     \\ \cmidrule{3-8}
                                               &                       & 1-shot              & 5-shot              & 1-shot              & 5-shot              & 1-shot              & 5-shot              \\ \midrule
MetaGAN \cite{metaGAN}                         & ConvNet-4             & 52.71\ci{0.64}      & 68.63\ci{0.67}      & --                  & --                  & --                  & --                  \\
$\Delta$-Encoder$\dagger$ \cite{deltaencoder}  & VGG-16                & 59.90\cip           & 69.70\cip           & 69.80\ci{0.46}      & 82.60\ci{0.35}      & 66.70\cip           & 79.80\cip           \\
SalNet \cite{salnet}                           & ResNet-101            & 62.22\ci{0.87}      & 77.95\ci{0.65}      & --                  & --                  & --                  & --                  \\
DTN \cite{DTN}                                 & Resnet-12             & 63.45\ci{0.86}      & 77.91\ci{0.62}      & 72.00\cip           & 85.10\cip           & 71.50\cip           & 82.80\cip           \\ \midrule
Dual TriNet \cite{semanticAugmentation}        & ResNet-18             & 58.80\ci{1.37}      & 76.71\ci{0.69}      & 69.61\cip           & 84.10\cip           & 63.41\ci{0.64}      & 78.43\ci{0.64}      \\
IDeMe-Net \cite{Ideme}                         & ResNet-18             & 59.14\ci{0.86}      & 74.63\ci{0.74}      & --                  & --                  & --                  & --                  \\
AFHN \cite{AFHN}                               & ResNet-18             & 62.38\ci{0.72}      & 78.16\ci{0.56}      & 70.53\ci{1.01}      & 83.95\ci{0.63}      & 68.32\ci{0.93}      & 81.45\ci{0.87}      \\
VI-Net \cite{VIFSL}                            & ResNet-18             & 61.05\cip           & 78.60\cip           & 74.76\cip           & 86.84\cip           & --                  & --                  \\ \midrule
Baseline~\eq{cls}                              & ResNet-18             & 56.81\ci{0.81}      & 78.31\ci{0.59}      & 67.14\ci{0.89}      & 86.22\ci{0.50}      & 65.71\ci{0.95}      & 84.68\ci{0.61}      \\
Baseline-KD~\eq{kl}                            & ResNet-18             & 59.62\ci{0.85}      & 79.31\ci{0.62}      & 70.85\ci{0.90}      & 87.64\ci{0.48}      & 69.15\ci{0.94}      & 85.89\ci{0.59}      \\
VFH (ours)                                     & ResNet-18             & 61.88\ci{0.85}      & 79.63\ci{0.61}      & 75.44\ci{0.85}      & 87.82\ci{0.47}      & 72.31\ci{0.91}      & 85.64\ci{0.64}      \\
\tb{TFH (ours)}                                & ResNet-18             & 64.49\ci{0.84}      & 79.94\ci{0.60}      & 75.66\ci{0.85}      & 88.39\ci{0.49}      & 73.77\ci{0.85}      & 86.68\ci{0.63}      \\
\tb{TFH-ft (ours)}                             & ResNet-18             & \tb{65.07}\ci{0.82} & \tb{80.81}\ci{0.61} & \tb{75.76}\ci{0.83} & \tb{88.60}\ci{0.47} & \tb{74.77}\ci{0.90} & \tb{86.88}\ci{0.59} \\
\bottomrule
\end{tabular}
\vspace{6pt}
\caption{\emph{Standard few-shot classification}. Comparison of our TFH, variants and baselines to SOTA few-shot data augmentation methods. Baseline~\eq{cls}, Baseline-KD~\eq{kl}: prototypical classifier at inference, no feature generation. VFH: our vector feature hallucinator; TFH: our tensor feature hallucinator; TFH-ft: our tensor feature hallucinator followed by novel-task fine-tuning. $\dagger$: Delta-encoder uses VGG-16 backbone for \emph{mini}ImageNet and CIFAR-FS and \emph{ResNet-18} for CUB.}
\label{tab:sota}
\end{table*}

\subsection{Setup}
\label{sec:setup}


\paragraph{Tasks}

We consider $N$-way, $K$-shot classification tasks with $N = 5$ randomly sampled novel classes and $K \in \{1, 5\}$ examples drawn at random per class as support set $S$, that is, $L = 5K$ examples in total. For the query set $Q$, we draw $15$ additional examples per class, that is, $75$ examples in total, which is the most common choice~\cite{tpn, learning2selftrain, transmatch}. We measure the \emph{classification accuracy} as the percentage of correctly classified queries per task. To reduce the variance, we report mean accuracy and 95$\%$ confidence interval over 600 tasks per experiment, similarly to~\cite{AFHN}.

\paragraph{Hyperparameters}

For ResNet-18, we generate $M=100$ and $M=2$ features per class in 1-shot and 5-shot tasks respectively using TFH and $M=100$ using TFH-ft. For ResNet-12, we generate $M=1000$ and $M=1$ features per class in 1-shot and 5-shot tasks respectively using TFH and $M=5$ using TFH-ft.

\paragraph{Baselines: no augmentation}

We define baselines consisting only of the embedding network $f_\theta$~\eq{cls} or $f_{\theta'}$~\eq{kl} at representation learning and a prototypical classifier at inference, without feature hallucination. We refer to them as Baseline~\eq{cls} and Baseline-KD~\eq{kl}, respectively.

\paragraph{Baseline: vector feature hallucinator (VHF)}

To validate the benefit of generating tensor features, we also generate \emph{vector} features by using $\bar{f}_{\theta'}: \cX \to \real^d$ including GAP~\eq{kl} as embedding model. In this case, the \emph{conditioner} $h: \real^d \to \real^{d'}$ consists of two fully-connected layers with a ReLU activation in-between. The \emph{generator} $g: \real^{k+d'} \to \real^d$ also consists of two fully-connected layers with a ReLU activation in-between and a sigmoid function at the end. The dimension $d'$ of the class-conditional vector as well as the dimensions of the hidden layers of both the conditioner and the generator are all set to 512.

\paragraph{Competitors}

We compare our method with state-of-the-art data augmentation methods for few-shot learning, including MetaGAN~\cite{metaGAN}, $\Delta$-encoder~\cite{deltaencoder}, \emph{salient network} (SalNet)~\cite{salnet}, \emph{diversity transfer network}~(DTN) \cite{DTN}, dual TriNet~\cite{semanticAugmentation}, \emph{image deformation meta-network} (IDeMe-Net)~\cite{Ideme}, \emph{adversarial feature hallucination network} (AFHN)~\cite{AFHN} and \emph{variational inference network} (VI-Net)~\cite{VIFSL}.


\subsection{Comparison with the state of the art}

\paragraph{Standard few-shot classification}

\autoref{tab:sota} compares our method with baselines and the state of the art. Most important are the comparisons with \cite{semanticAugmentation, Ideme, AFHN, VIFSL}, which use the same backbone. Our TFH provides new state of the art in all datasets in both 1-shot and 5-shot tasks, outperforming all competing few-shot data augmentation methods.

TFH is superior to VFH, especially in \emph{mini}ImageNet 1-shot, providing almost 3\% performance improvement. The importance of tensor features is clear from the fact that VHF is worse than AFHN~\cite{AFHN} while TFH is better than AFHN by more than 2\% in \emph{mini}ImageNet 1-shot. VFH still outperforms the state of the art in all other experiments, highlighting the effectiveness of our new loss function. Novel-task \emph{fine-tuning} (TFH-ft) is mostly beneficial, impressively even in 1-shot tasks. This shows that our tensor hallucinator is robust and avoids overfitting. \emph{Self-distillation} provides a significant gain in all experiments.

\begin{table}
\small
\centering
\setlength\tabcolsep{4pt}
\begin{tabular}{lcccccc} \toprule
\mr{2}{\Th{Method}} & \mc{2}{\Th{\emph{m}IN}$\to$\Th{CUB}}      & \mc{2}{\Th{\emph{m}IN}$\to$\Th{CIFAR-FS}}  \\ \cmidrule{2-5}
                    & 1-shot              & 5-shot              & 1-shot               & 5-shot              \\ \midrule
Baseline~\eq{cls}   & 43.14\ci{0.78}      & 62.20\ci{0.70}      & 50.25\ci{0.86}       & 69.43\ci{0.74}      \\
Baseline-KD~\eq{kl} & 44.40\ci{0.82}      & \tb{63.83}\ci{0.73} & 51.54\ci{0.89}       & 70.40\ci{0.72}      \\
\tb{VFH (ours)}     & 44.77\ci{0.79}      & 62.61\ci{0.73}      & 50.36\ci{0.87}       & 69.31\ci{0.74}      \\
\tb{TFH (ours)}     & 45.67\ci{0.80}      & 63.08\ci{0.73}      & 51.82\ci{0.89}       & 69.77\ci{0.76}      \\
\tb{TFH-ft (ours)}  & \tb{45.96}\ci{0.80} & 63.64\ci{0.74}      &  \tb{53.07}\ci{0.86} & \tb{71.29}\ci{0.75} \\
\bottomrule
\end{tabular}
\vspace{6pt}
\caption{\emph{Few-shot cross-domain classification}. Comparison of our TFH to variants and baselines as defined in \autoref{tab:sota}, using ResNet-18 trained on \emph{mini}ImageNet (\emph{m}IN).}
\label{tab:cross_domain}
\end{table}

\paragraph{Cross-domain few-shot classification}

We investigate the ability of our tensor hallucinator to address domain shift, carrying out experiments in the \emph{cross-domain} few-shot classification setting proposed by \cite{closerlook}. We are not aware of any other synthetic data generation method that has been applied to this setting. We train the ResNet-18 backbone and our tensor hallucinator on \emph{mini}ImageNet as $D_{\base}$ and solve novel tasks on CUB and CIFAR-FS as $D_{\novel}$. As shown in \autoref{tab:cross_domain}, our tensor hallucinator can address the domain shift effectively, especially in 1-shot tasks, even without fine-tuning. Even though the results in 5-shot tasks are less impressive, our hallucinator still outperforms the Baseline-KD on \emph{mini}ImageNet$\to$CIFAR-FS, while being on par on \emph{mini}ImageNet$\to$CUB.

\begin{table}
\small
\centering
\setlength\tabcolsep{4pt}
\begin{tabular}{lcccccc} \toprule
\Th{Method}                                     & \mc{2}{\Th{\emph{mini}ImageNet}}          & \mc{2}{\Th{CUB}}                          \\ \cmidrule{2-5}
                                                & 1-shot              & 5-shot              & 1-shot              & 5-shot              \\ \midrule
\mc{5}{\Th{Logistic regression}}\\ \midrule
Baseline~\eq{cls}                               & 59.20\ci{0.82}      & 77.71\ci{0.61}      & 69.44\ci{0.86}      & 86.19\ci{0.49}      \\
Baseline-KD~\eq{kl}                             & 61.83\ci{0.82}      & 79.27\ci{0.61}      & 72.74\ci{0.88}      & 87.71\ci{0.49}      \\
VFH (ours)                                      & 62.37\ci{0.83}      & 79.70\ci{0.59}      & 75.06\ci{0.88}      & 87.95\ci{0.46}      \\
\tb{TFH (ours)}                                 & 64.03\ci{0.84}      & 79.93\ci{0.60}      & 75.08\ci{0.85}      & \tb{88.82}\ci{0.46} \\
\tb{TFH-ft (ours)}                              & \tb{64.83}\ci{0.82} & \tb{80.49}\ci{0.61} & \tb{75.43}\ci{0.85} & 88.52\ci{0.48}      \\ \midrule
\mc{5}{\Th{Support vector machine}}\\ \midrule
Baseline~\eq{cls}                               & 57.12\ci{0.84}      & 76.45\ci{0.62}      & 67.24\ci{0.87}      & 84.72\ci{0.52}      \\
Baseline-KD~\eq{kl}                             & 60.21\ci{0.84}      & 78.28\ci{0.61}      & 71.23\ci{0.89}      & 86.34\ci{0.52}      \\
VFH (ours)                                      & 61.95\ci{0.85}      & 79.15\ci{0.60}      & 75.43\ci{0.85}      & 87.46\ci{0.47}      \\
\tb{TFH (ours)}                                 & 64.20\ci{0.83}      & 79.62\ci{0.60}      & 75.64\ci{0.85}      & \tb{88.74}\ci{0.45} \\
\tb{TFH-ft (ours)}                              & \tb{65.06}\ci{0.82} & \tb{80.33}\ci{0.60} & \tb{75.77}\ci{0.83} & 88.22\ci{0.46}      \\
\bottomrule
\end{tabular}
\vspace{6pt}
\caption{\emph{Alternative classifiers}. Our TFH, variants and baselines as defined in \autoref{tab:sota}, using ResNet-18, where at inference, the prototypical classifier is replaced by logistic regression or SVM.}
\label{tab:dif_class}
\end{table}

\subsection{Ablations}
\label{sec:ablation}

\paragraph{Alternative classifiers}

We investigate the effect of replacing the prototypical classifier~\cite{prototypical} by alternative classifiers at inference, applied to the same augmented support set according the features generated by our tensor hallucinator. We consider \emph{logistic regression} and \emph{support vector machine} (SVM) classifiers, both using the scikit-learn framework~\cite{sklearn}. As shown in \autoref{tab:dif_class}, our tensor hallucinator clearly provides the best accuracy results irrespective of the classifier, while the performance of all three classifiers is on par. This indicates that individual generated features are also useful, not just the centroid per class.

\begin{table}
\small
\centering
\setlength\tabcolsep{4pt}
\begin{tabular}{lcccccc} \toprule
\mr{2}{\Th{Method}}     & \mc{2}{\Th{\emph{mini}ImageNet}}          & \mc{2}{\Th{CUB}}                          \\
\cmidrule{2-5}
                        & 1-shot              & 5-shot              & 1-shot              & 5-shot              \\
\midrule
\mc{5}{\Th{Prototypical classifier}} \\ \midrule
Baseline                & 59.94\ci{0.84}      & 78.65\ci{0.61}       & 66.72\ci{0.90}      & 84.27\ci{0.59}      \\
\tb{TFH (ours)}         & 64.69\ci{0.83}      & 79.58\ci{0.59}       & 68.71\ci{0.90}      & 84.83\ci{0.53}      \\
\tb{TFH-ft (ours)}      & \tb{65.16}\ci{0.86} & \tb{79.83}\ci{0.65} & \tb{70.52}\ci{0.87} & \tb{85.02}\ci{0.55}\\
\midrule
\mc{5}{\Th{Logistic regression}} \\ \midrule
Baseline                & 61.83\ci{0.85}      & 79.60\ci{0.60}       & 68.49\ci{0.49}      & 84.75\ci{0.59}      \\
\tb{TFH (ours)}         & 64.20\ci{0.84}      & 80.21\ci{0.58}      & 68.37\ci{0.92}      & \tb{85.12}\ci{0.55}      \\
\tb{TFH-ft (ours)}      & \tb{64.53}\ci{0.53} & \tb{80.25}\ci{0.62} & \tb{69.59}\ci{0.88} & 85.08\ci{0.55} \\
\midrule
\mc{5}{\Th{Support vector machine}} \\ \midrule
Baseline                & 59.94\ci{0.84}      & 78.43\ci{0.60}      & 66.72\ci{0.90}      & 83.29\ci{0.60}      \\
\tb{TFH(ours)}          & 64.62\ci{0.83}      & 79.56\ci{0.58}      & 68.75\ci{0.90}      & 84.05\ci{0.56} \\
\tb{TFH-ft(ours)}       & \tb{64.64}\ci{0.86} & \tb{79.88}\ci{0.62} & \tb{69.42}\ci{0.88} & \tb{84.41}\ci{0.55}     \\
\bottomrule
\end{tabular}
\vspace{6pt}
\caption{\emph{Alternative backbone network}. Our TFH, variants and baselines as defined in \autoref{tab:sota}, using the publicly available pre-trained ResNet-12 backbones provided by~\cite{deepEMD}.}
\label{tab:res12}
\end{table}


\paragraph{Alternative backbone network}

To investigate the effect of using alternative backbone networks, we replace ResNet-18 by ResNet-12 and use the pre-trained networks on \emph{mini}ImageNet and CUB  provided by DeepEMD~\cite{deepEMD}\footnote{\url{https://github.com/icoz69/DeepEMD}}. We use all three classifiers: prototypical, logistic regression and SVM. As shown in \autoref{tab:res12}, our tensor hallucinator provides the best results in all settings with a significant performance gain of around 3-5\% in 1-shot tasks in prototypical and support vector machines classifiers in both datasets.

\begin{table}
\small
\centering
\setlength\tabcolsep{4pt}
\begin{tabular}{ccccccc} \toprule
\mr{2}{\Th{\# Feat. $M$}} & \mc{2}{\Th{ResNet-18}}                    & \mc{2}{\Th{ResNet-12}}                    \\ \cmidrule{2-5}
                          & 1-shot              & 5-shot              & 1-shot              & 5-shot              \\ \midrule
0                         & 59.62\ci{0.85}      & 79.31\ci{0.62}      & 59.94\ci{0.84}      & 78.65\ci{0.61}      \\
1                         & 62.43\ci{0.83}      & 79.33\ci{0.62}      & 63.88\ci{0.85} & \tb{79.58}\ci{0.59}\\
2                         & 63.57\ci{0.87}      & \tb{79.94}\ci{0.60} & 64.01\ci{0.83} & 79.44\ci{0.61}\\ 
5                         & 64.42\ci{0.83}      & 79.77\ci{0.61}   & 64.12\ci{0.83} & 79.49\ci{0.61}\\   
10                        & 63.90\ci{0.81}      & 79.64\ci{0.60}     & 64.01\ci{0.85} & 79.56\ci{0.63}\\ 
50                        & 63.44\ci{0.86}      & 79.43\ci{0.63}     & 63.67\ci{0.84} & 79.18\ci{0.65}\\ 
100                       & \tb{64.49}\ci{0.84} & 79.27\ci{0.60}    & 63.66\ci{0.86} & 78.89\ci{0.65}\\  
500                       & 64.25\ci{0.85}      & 79.71\ci{0.62}  & 64.11\ci{0.86} & 79.38\ci{0.65}\\    
1000                      & 63.76\ci{0.87}      & 79.52\ci{0.62}    & \tb{64.69}\ci{0.83} & 79.48\ci{0.60}\\ \bottomrule  
\end{tabular}
\vspace{6pt}
\caption{\emph{Effect of number of generated features, $M$}. Using our tensor feature hallucinator (TFH) with the default prototypical classifier and without fine-tuning on \emph{mini}ImageNet.}
\label{tab:gen_feat}
\end{table}

\paragraph{Effect of number of generated features}

We investigate the effect of the number of generated features per class, $M$, in \autoref{tab:gen_feat}. The results are similar for both backbones, highlighting the generality of our TFH. In all settings, regardless of number of generated tensor features, our hallucinator provides better performance. Remarkably, even 1 or 2 generated features per class provide a significant gain of around 4\% in 1-shot tasks for both networks when compared to 0 (Baseline-KD for ResNet-18 and Baseline for ResNet-12). Interestingly, the performance of our hallucinator does not degrade even when 1000 tensor features are generated, still outperforming the baseline models and providing the best 1-shot accuracy for ResNet-12. This is significant, since when generating 1000 features, the vast majority of the support set consists only of synthetic data.




\begin{figure}[h]
\small
\centering
\setlength\tabcolsep{1pt}
\begin{tabular}{cc}
\fig[0.5]{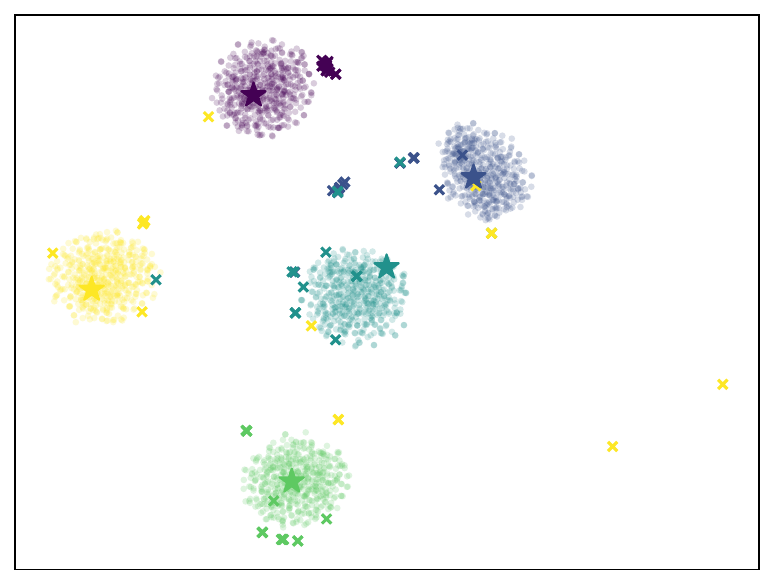} &
\fig[0.5]{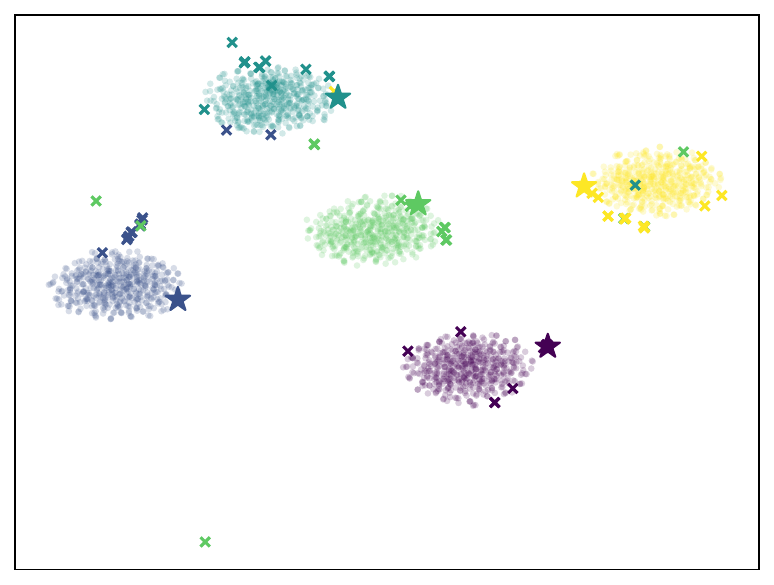} \\
(a) \emph{mini}ImageNet, ResNet-18 & (b) CUB, ResNet-18 \\[3pt]
\fig[0.5]{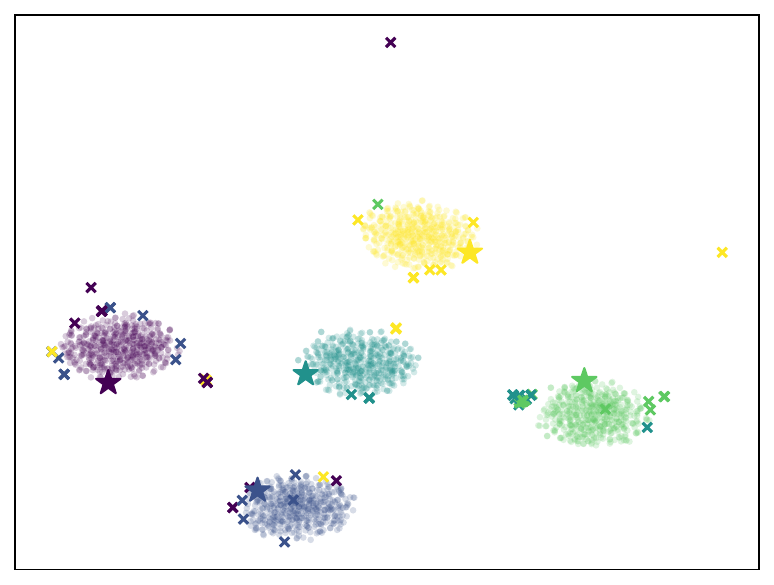} &
\fig[0.5]{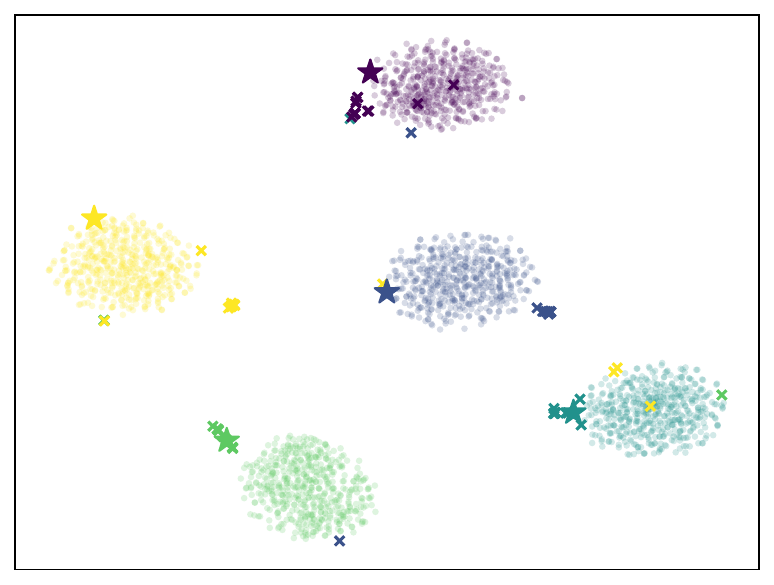} \\
(c) \emph{mini}ImageNet, ResNet-12 & (d) CUB, ResNet-12 \\
\end{tabular}
\vspace{3pt}
\caption{t-SNE visualization of the augmented support feature set of an 1-shot task using both ResNet-18 and ResNet-12 backbones on \emph{mini}ImageNet and CUB. Colors indicate different classes. $\star$: support features; $\bullet$: generated features; `$\times$': query features.}
\label{fig:tsne}
\end{figure}
\paragraph{Visualization of feature embeddings}

To investigate the generated feature distribution, we sample a 5-way, 1-shot novel task, use the support set to generate 500 novel features per class and visualize the augmented support set in 2D using t-SNE~\cite{tsne}. As shown in \autoref{fig:tsne}, the generated features of each class are clustered together, with distinct class boundaries. This synthetic class-conditional distribution is improving the classifier performance in the absence of true data. It can be seen that the query examples are much more scattered, showing that the variance of each novel class can be large, highlighting the inherent difficulty of the few-shot classification problem. If only the support examples were used, making predictions would be much harder because of how scattered the queries are.

\section{Conclusion}
\label{conclusion}

In this work, we have introduced a conceptually simple tensor feature hallucinator that improves the state of the art on synthetic data generation for few-shot learning classification. We have provided evidence showing that the structural properties of tensors provide a significant performance gain, allowing for a simplification of the loss function and training regime.
We have also shown the importance of complementing with improved representation learning, as well as task adaptation by fine-tuning on the augmented support set, which reduces the risk of overfitting.

Potential future directions include: improving our hallucinator architecture; 
experimenting with different loss functions to train our hallucinator; and investigating the use of our hallucinator in different settings, such as long-tailed recognition or incremental learning.

{\small
\bibliographystyle{ieee_fullname}
\bibliography{egbib}

\begin{thebibliography}{10}\itemsep=-1pt

\bibitem{wGAN}
Martin Arjovsky, Soumith Chintala, and L{\'e}on Bottou.
\newblock Wasserstein generative adversarial networks.
\newblock In {\em ICML}, 2017.

\bibitem{bertinetto}
Luca Bertinetto, Joao~F Henriques, Philip~HS Torr, and Andrea Vedaldi.
\newblock Meta-learning with differentiable closed-form solvers.
\newblock {\em arXiv preprint arXiv:1805.08136}, 2018.

\bibitem{biggan}
Andrew Brock, Jeff Donahue, and Karen Simonyan.
\newblock Large scale {GAN} training for high fidelity natural image synthesis.
\newblock {\em arXiv preprint arXiv:1809.11096}, 2018.

\bibitem{DTN}
Mengting Chen, Yuxin Fang, Xinggang Wang, Heng Luo, Yifeng Geng, Xinyu Zhang,
  Chang Huang, Wenyu Liu, and Bo Wang.
\newblock Diversity transfer network for few-shot learning.
\newblock In {\em AAAI}, 2020.

\bibitem{chen2020simple}
Ting Chen, Simon Kornblith, Mohammad Norouzi, and Geoffrey Hinton.
\newblock A simple framework for contrastive learning of visual
  representations.
\newblock In {\em ICML}, 2020.

\bibitem{closerlook}
Wei-Yu Chen, Yen-Cheng Liu, Zsolt Kira, Yu-Chiang Wang, and Jia-Bin Huang.
\newblock A closer look at few-shot classification.
\newblock In {\em ICLR}, 2019.

\bibitem{Ideme}
Zitian Chen, Yanwei Fu, Yu-Xiong Wang, Lin Ma, Wei Liu, and Martial Hebert.
\newblock Image deformation meta-networks for one-shot learning.
\newblock In {\em CVPR}, 2019.

\bibitem{semanticAugmentation}
Z. {Chen}, Y. {Fu}, Y. {Zhang}, Y. {Jiang}, X. {Xue}, and L. {Sigal}.
\newblock Multi-level semantic feature augmentation for one-shot learning.
\newblock {\em IEEE Transactions on Image Processing}, 2019.

\bibitem{donahue2014decaf}
Jeff Donahue, Yangqing Jia, Oriol Vinyals, Judy Hoffman, Ning Zhang, Eric
  Tzeng, and Trevor Darrell.
\newblock Decaf: A deep convolutional activation feature for generic visual
  recognition.
\newblock In {\em ICML}, 2014.

\bibitem{MAML}
Chelsea Finn, Pieter Abbeel, and Sergey Levine.
\newblock Model-agnostic meta-learning for fast adaptation of deep networks.
\newblock In {\em ICML}, 2017.

\bibitem{ganin2015unsupervised}
Yaroslav Ganin and Victor Lempitsky.
\newblock Unsupervised domain adaptation by backpropagation.
\newblock In {\em ICML}, 2015.

\bibitem{fewshotGNN1}
Victor Garcia and Joan Bruna.
\newblock Few-shot learning with graph neural networks.
\newblock {\em arXiv preprint arXiv:1711.04043}, 2017.

\bibitem{gidarisFSL}
Spyros Gidaris and Nikos Komodakis.
\newblock Dynamic few-shot visual learning without forgetting.
\newblock In {\em CVPR}, 2018.

\bibitem{gidarisRot}
Spyros Gidaris, Praveer Singh, and Nikos Komodakis.
\newblock Unsupervised representation learning by predicting image rotations.
\newblock In {\em ICLR}, 2018.

\bibitem{GANSgoodfellow}
Ian~J Goodfellow, Jean Pouget-Abadie, Mehdi Mirza, Bing Xu, David Warde-Farley,
  Sherjil Ozair, Aaron Courville, and Yoshua Bengio.
\newblock Generative adversarial networks.
\newblock {\em arXiv preprint arXiv:1406.2661}, 2014.

\bibitem{lowshot2}
Bharath Hariharan and Ross Girshick.
\newblock Low-shot visual recognition by shrinking and hallucinating features.
\newblock In {\em CVPR}, 2017.

\bibitem{resnet18}
Kaiming He, Xiangyu Zhang, Shaoqing Ren, and Jian Sun.
\newblock Deep residual learning for image recognition.
\newblock In {\em CVPR}, 2016.

\bibitem{FID}
Martin Heusel, Hubert Ramsauer, Thomas Unterthiner, Bernhard Nessler, and Sepp
  Hochreiter.
\newblock {GANs} trained by a two time-scale update rule converge to a local
  nash equilibrium.
\newblock In {\em NeurIPS}, 2017.

\bibitem{fewshotCUB}
Nathan Hilliard, Lawrence Phillips, Scott Howland, Art{\"e}m Yankov, Courtney~D
  Corley, and Nathan~O Hodas.
\newblock Few-shot learning with metric-agnostic conditional embeddings.
\newblock {\em arXiv preprint arXiv:1802.04376}, 2018.

\bibitem{flow++}
Jonathan Ho, Xi Chen, Aravind Srinivas, Yan Duan, and Pieter Abbeel.
\newblock Flow++: Improving flow-based generative models with variational
  dequantization and architecture design.
\newblock In {\em ICML}, 2019.

\bibitem{pggan}
Tero Karras, Timo Aila, Samuli Laine, and Jaakko Lehtinen.
\newblock Progressive growing of {GANs} for improved quality, stability, and
  variation.
\newblock {\em arXiv preprint arXiv:1710.10196}, 2017.

\bibitem{styleganADA}
Tero Karras, Miika Aittala, Janne Hellsten, Samuli Laine, Jaakko Lehtinen, and
  Timo Aila.
\newblock Training generative adversarial networks with limited data.
\newblock {\em arXiv preprint arXiv:2006.06676}, 2020.

\bibitem{stylegan}
Tero Karras, Samuli Laine, and Timo Aila.
\newblock A style-based generator architecture for generative adversarial
  networks.
\newblock In {\em CVPR}, 2019.

\bibitem{fewshotGNN2}
Jongmin Kim, Taesup Kim, Sungwoong Kim, and Chang~D Yoo.
\newblock Edge-labeling graph neural network for few-shot learning.
\newblock In {\em CVPR}, 2019.

\bibitem{glow}
Diederik~P Kingma and Prafulla Dhariwal.
\newblock Glow: Generative flow with invertible 1x1 convolutions.
\newblock {\em arXiv preprint arXiv:1807.03039}, 2018.

\bibitem{VAE}
Diederik~P Kingma and Max Welling.
\newblock Auto-encoding variational bayes.
\newblock {\em arXiv preprint arXiv:1312.6114}, 2013.

\bibitem{siamese}
Gregory Koch, Richard Zemel, and Ruslan Salakhutdinov.
\newblock Siamese neural networks for one-shot image recognition.
\newblock In {\em ICML workshop}, 2015.

\bibitem{ECCV2020_211}
Alexander Kolesnikov, Lucas Beyer, Xiaohua Zhai, Joan Puigcerver, Jessica Yung,
  Sylvain Gelly, and Neil Houlsby.
\newblock Big transfer ({BiT}): General visual representation learning.
\newblock In {\em ECCV}, 2020.

\bibitem{cifar100db}
Alex Krizhevsky, Geoffrey Hinton, et~al.
\newblock Learning multiple layers of features from tiny images.
\newblock Technical report, 2009.

\bibitem{imagenet}
Alex Krizhevsky, Ilya Sutskever, and Geoffrey~E Hinton.
\newblock Imagenet classification with deep convolutional neural networks.
\newblock In {\em NeurIPS}, 2012.

\bibitem{ilpc}
Michalis Lazarou, Yannis Avrithis, and Tania Stathaki.
\newblock Iterative label cleaning for transductive and semi-supervised
  few-shot learning.
\newblock {\em arXiv preprint arXiv:2012.07962}, 2020.

\bibitem{metaridge}
Kwonjoon Lee, Subhransu Maji, Avinash Ravichandran, and Stefano Soatto.
\newblock Meta-learning with differentiable convex optimization.
\newblock In {\em CVPR}, 2019.

\bibitem{categorytraversalFSL}
Hongyang Li, David Eigen, Samuel Dodge, Matthew Zeiler, and Xiaogang Wang.
\newblock Finding task-relevant features for few-shot learning by category
  traversal.
\newblock In {\em CVPR}, 2019.

\bibitem{AFHN}
Kai Li, Yulun Zhang, Kunpeng Li, and Yun Fu.
\newblock Adversarial feature hallucination networks for few-shot learning.
\newblock In {\em CVPR}, 2020.

\bibitem{learning2selftrain}
Xinzhe Li, Qianru Sun, Yaoyao Liu, Qin Zhou, Shibao Zheng, Tat-Seng Chua, and
  Bernt Schiele.
\newblock Learning to self-train for semi-supervised few-shot classification.
\newblock In {\em NeurIPS}, 2019.

\bibitem{denseclassification}
Yann Lifchitz, Yannis Avrithis, Sylvaine Picard, and Andrei Bursuc.
\newblock Dense classification and implanting for few-shot learning.
\newblock In {\em CVPR}, 2019.

\bibitem{few-shotGAN}
Bingchen Liu, Yizhe Zhu, Kunpeng Song, and Ahmed Elgammal.
\newblock Towards faster and stabilized {GAN} training for high-fidelity
  few-shot image synthesis.
\newblock {\em arXiv e-prints}, pages arXiv--2101, 2021.

\bibitem{funit}
Ming-Yu Liu, Xun Huang, Arun Mallya, Tero Karras, Timo Aila, Jaakko Lehtinen,
  and Jan Kautz.
\newblock Few-shot unsupervised image-to-image translation.
\newblock In {\em CVPR}, 2019.

\bibitem{tpn}
Yanbin Liu, Juho Lee, Minseop Park, Saehoon Kim, Eunho Yang, Sung~Ju Hwang, and
  Yi Yang.
\newblock Learning to propagate labels: Transductive propagation network for
  few-shot learning.
\newblock {\em arXiv preprint arXiv:1805.10002}, 2018.

\bibitem{VIFSL}
Qinxuan Luo, Lingfeng Wang, Jingguo Lv, Shiming Xiang, and Chunhong Pan.
\newblock Few-shot learning via feature hallucination with variational
  inference.
\newblock In {\em WACV}, 2021.

\bibitem{manifoldmixup}
Puneet Mangla, Nupur Kumari, Abhishek Sinha, Mayank Singh, Balaji
  Krishnamurthy, and Vineeth~N Balasubramanian.
\newblock Charting the right manifold: Manifold mixup for few-shot learning.
\newblock In {\em WACV}, 2020.

\bibitem{language}
Tom{\'a}{\v{s}} Mikolov, Martin Karafi{\'a}t, Luk{\'a}{\v{s}} Burget, Jan
  {\v{C}}ernock{\`y}, and Sanjeev Khudanpur.
\newblock Recurrent neural network based language model.
\newblock In {\em Eleventh annual conference of the international speech
  communication association}, 2010.

\bibitem{metaNet}
Tsendsuren Munkhdalai and Hong Yu.
\newblock Meta networks.
\newblock In {\em ICML}, 2017.

\bibitem{reptile}
Alex Nichol, Joshua Achiam, and John Schulman.
\newblock On first-order meta-learning algorithms.
\newblock {\em arXiv preprint arXiv:1803.02999}, 2018.

\bibitem{oh2016deep}
Hyun Oh~Song, Yu Xiang, Stefanie Jegelka, and Silvio Savarese.
\newblock Deep metric learning via lifted structured feature embedding.
\newblock In {\em CVPR}, 2016.

\bibitem{condpixelCNN}
Aaron van~den Oord, Nal Kalchbrenner, Oriol Vinyals, Lasse Espeholt, Alex
  Graves, and Koray Kavukcuoglu.
\newblock Conditional image generation with pixelcnn decoders.
\newblock {\em arXiv preprint arXiv:1606.05328}, 2016.

\bibitem{tadam}
Boris Oreshkin, Pau~Rodr{\'\i}guez L{\'o}pez, and Alexandre Lacoste.
\newblock {TADAM}: Task dependent adaptive metric for improved few-shot
  learning.
\newblock In {\em NeurIPS}, 2018.

\bibitem{sklearn}
F. Pedregosa, G. Varoquaux, A. Gramfort, V. Michel, B. Thirion, O. Grisel, M.
  Blondel, P. Prettenhofer, R. Weiss, V. Dubourg, J. Vanderplas, A. Passos, D.
  Cournapeau, M. Brucher, M. Perrot, and E. Duchesnay.
\newblock Scikit-learn: Machine learning in {P}ython.
\newblock {\em JMLR}, 2011.

\bibitem{imprintedweights}
Hang Qi, Matthew Brown, and David~G Lowe.
\newblock Low-shot learning with imprinted weights.
\newblock In {\em CVPR}, 2018.

\bibitem{ravilstm}
Sachin Ravi and Hugo Larochelle.
\newblock Optimization as a model for few-shot learning.
\newblock 2016.

\bibitem{rebuffi2017learning}
Sylvestre-Alvise Rebuffi, Hakan Bilen, and Andrea Vedaldi.
\newblock Learning multiple visual domains with residual adapters.
\newblock {\em arXiv preprint arXiv:1705.08045}, 2017.

\bibitem{rebuffi2017icarl}
Sylvestre-Alvise Rebuffi, Alexander Kolesnikov, Georg Sperl, and Christoph~H
  Lampert.
\newblock {iCaRL}: Incremental classifier and representation learning.
\newblock In {\em CVPR}, 2017.

\bibitem{embeddingpropagation}
Pau Rodríguez, Issam Laradji, Alexandre Drouin, and Alexandre Lacoste.
\newblock Embedding propagation: Smoother manifold for few-shot classification.
\newblock {\em ECCV}, 2020.

\bibitem{IS}
Tim Salimans, Ian Goodfellow, Wojciech Zaremba, Vicki Cheung, Alec Radford, Xi
  Chen, and Xi Chen.
\newblock Improved techniques for training {GANs}.
\newblock In {\em NIPS}, 2016.

\bibitem{santoro}
Adam Santoro, Sergey Bartunov, Matthew Botvinick, Daan Wierstra, and Timothy
  Lillicrap.
\newblock Meta-learning with memory-augmented neural networks.
\newblock In {\em ICML}, 2016.

\bibitem{schmidhuber1987evolutionary}
J{\"u}rgen Schmidhuber.
\newblock Evolutionary principles in self-referential learning, or on learning
  how to learn: the meta-meta-... hook.
\newblock Master's thesis, Technische Universit{\"a}t M{\"u}nchen, 1987.

\bibitem{deltaencoder}
Eli Schwartz, Leonid Karlinsky, Joseph Shtok, Sivan Harary, Mattias Marder,
  Abhishek Kumar, Rogerio Feris, Raja Giryes, and Alex Bronstein.
\newblock Delta-encoder: an effective sample synthesis method for few-shot
  object recognition.
\newblock In {\em NeurIPS}, 2018.

\bibitem{prototypical}
Jake Snell, Kevin Swersky, and Richard Zemel.
\newblock Prototypical networks for few-shot learning.
\newblock In {\em NeurIPS}, 2017.

\bibitem{sohn2020fixmatch}
Kihyuk Sohn, David Berthelot, Chun-Liang Li, Zizhao Zhang, Nicholas Carlini,
  Ekin~D Cubuk, Alex Kurakin, Han Zhang, and Colin Raffel.
\newblock Fixmatch: Simplifying semi-supervised learning with consistency and
  confidence.
\newblock {\em arXiv preprint arXiv:2001.07685}, 2020.

\bibitem{relationnet}
Flood Sung, Yongxin Yang, Li Zhang, Tao Xiang, Philip~HS Torr, and Timothy~M
  Hospedales.
\newblock Learning to compare: Relation network for few-shot learning.
\newblock In {\em CVPR}, 2018.

\bibitem{rfs}
Yonglong Tian, Yue Wang, Dilip Krishnan, Joshua~B Tenenbaum, and Phillip Isola.
\newblock Rethinking few-shot image classification: a good embedding is all you
  need?
\newblock {\em arXiv preprint arXiv:2003.11539}, 2020.

\bibitem{tsne}
Laurens Van~der Maaten and Geoffrey Hinton.
\newblock Visualizing data using {t-SNE}.
\newblock {\em JMLR}, 9(11), 2008.

\bibitem{pixelCNN}
Aaron Van~Oord, Nal Kalchbrenner, and Koray Kavukcuoglu.
\newblock Pixel recurrent neural networks.
\newblock In {\em ICML}, 2016.

\bibitem{vilalta2002perspective}
Ricardo Vilalta and Youssef Drissi.
\newblock A perspective view and survey of meta-learning.
\newblock {\em Artificial Intelligence Review}, 18(2):77--95, 2002.

\bibitem{matchingNets}
Oriol Vinyals, Charles Blundell, Timothy Lillicrap, Daan Wierstra, et~al.
\newblock Matching networks for one shot learning.
\newblock In {\em NIPS}, 2016.

\bibitem{frn_hariharan}
Davis Wertheimer, Luming Tang, and Bharath Hariharan.
\newblock Few-shot classification with feature map reconstruction networks.
\newblock In {\em CVPR}, 2021.

\bibitem{feat}
Han-Jia Ye, Hexiang Hu, De-Chuan Zhan, and Fei Sha.
\newblock Few-shot learning via embedding adaptation with set-to-set functions.
\newblock In {\em CVPR}, 2020.

\bibitem{tapnet}
Sung~Whan Yoon, Jun Seo, and Jaekyun Moon.
\newblock Tapnet: Neural network augmented with task-adaptive projection for
  few-shot learning.
\newblock In {\em ICML}, 2019.

\bibitem{transmatch}
Zhongjie Yu, L. Chen, Zhongwei Cheng, and Jiebo Luo.
\newblock {TransMatch}: A transfer-learning scheme for semi-supervised few-shot
  learning.
\newblock {\em CVPR}, 2020.

\bibitem{deepEMD}
Chi Zhang, Yujun Cai, Guosheng Lin, and Chunhua Shen.
\newblock {DeepEMD}: Few-shot image classification with differentiable earth
  mover's distance and structured classifiers.
\newblock In {\em CVPR}, 2020.

\bibitem{salnetFSL}
Hongguang Zhang, Jing Zhang, and Piotr Koniusz.
\newblock Few-shot learning via saliency-guided hallucination of samples.
\newblock In {\em CVPR}, 2019.

\bibitem{salnet}
Hongguang Zhang, Jing Zhang, and Piotr Koniusz.
\newblock Few-shot learning via saliency-guided hallucination of samples.
\newblock In {\em CVPR}, 2019.

\bibitem{metaGAN}
Ruixiang Zhang, Tong Che, Zoubin Ghahramani, Yoshua Bengio, and Yangqiu Song.
\newblock {MetaGAN}: An adversarial approach to few-shot learning.
\newblock {\em NeurIPS}, 2018.

\bibitem{CAVIA}
Luisa Zintgraf, Kyriacos Shiarli, Vitaly Kurin, Katja Hofmann, and Shimon
  Whiteson.
\newblock Fast context adaptation via meta-learning.
\newblock In {\em ICML}, 2019.

\end{thebibliography}
}

\clearpage



\textbf{\Large Supplementary material}

\appendix
\section{Datasets}
\label{sec:data}

\paragraph{\emph{mini}ImageNet}

This is a widely used few-shot image classification dataset~\cite{matchingNets, ravilstm}. It contains 100 randomly sampled classes from ImageNet~\cite{imagenet}. These 100 classes are split into 64 training (base) classes, 16 validation (novel) classes and 20 test (novel) classes. Each class contains 600 examples (images). We follow the commonly used split provided by~\cite{ravilstm}.

\paragraph{CUB}

This is a fine-grained image classification dataset consisting of 200 classes, each corresponding to a bird species. We follow the split defined by~\cite{closerlook, fewshotCUB}, with 100 training, 50 validation and 50 test classes.

\paragraph{CIFAR-FS}

This dataset is derived from CIFAR-100~\cite{cifar100db}, consisting of 100 classes with 600 examples per class. We follow the split provided by~\cite{closerlook}, with 64 training, 16 validation and 20 test classes.

When using ResNet-18 as a backbone network, images are resized to $224 \times 224$ for all datasets, similarly to other data augmentation methods \cite{AFHN, semanticAugmentation, Ideme, VIFSL}. When using ResNet-12, they are resized to $84 \times 84$, similarly to~\cite{deepEMD}.

\begin{table}[h]
\small
\centering
\setlength\tabcolsep{4pt}
\begin{tabular}{lc} \toprule
\mc{2}{\Th{Reconstructor network}} \\ \midrule
Layer                   & Output shape    \\ \midrule
Input                   & $512 \times 7 \times 7   $ \\
ResBlockA               & $256 \times 14 \times 14 $ \\
ResBlockA               & $128 \times 28 \times 28 $ \\
ResBlockA               & $64 \times 56 \times 56  $ \\
TranspConv3x3, stride=2 & $64 \times 113 \times 113$ \\
ResBlockB               & $3 \times 226 \times 224  $ \\
Bilinear interpolation  & $3 \times 224 \times 224 $ \\ \bottomrule
\end{tabular}
\vspace{6pt}
\caption{\emph{Image reconstructor architecture}. ResBlockA is exactly the same as ResBlockB except that it uses ReLU activation function, while ResBlockB uses sigmoid.}
\label{tab:recon_arch}
\end{table}


\section{Image reconstructors}

We carried out an experiment to investigate whether the output tensor features without global average pooling (GAP) can provide more spatial information to aid the reconstruction of the original image, when compared to vector features obtained by GAP. A similar experiment has been carried out by \cite{frn_hariharan} to visualize the tensor feature maps. We train two image reconstructors using a variant of an inverted ResNet-18 architecture with an additional transposed convolution layer, as shown in \autoref{tab:recon_arch}. The first is a tensor reconstructor, exactly as in \autoref{tab:recon_arch}. The second is a vector reconstructor taking a $512 \times 1 \times 1$ input. It is identical, except that it begins with an additional upsampling layer to adapt spatial resolution to $7 \times 7$.

\begin{figure*}[ht]
\small
\centering
\setlength\tabcolsep{1pt}
\begin{tabular}{c}
\fig{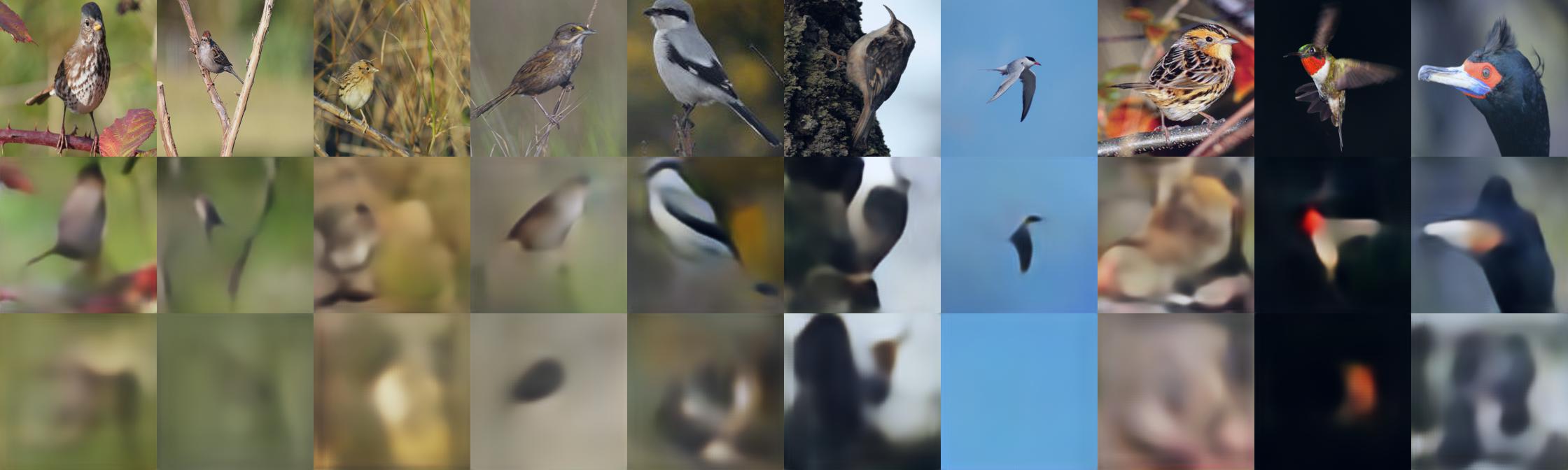}\\
\fig{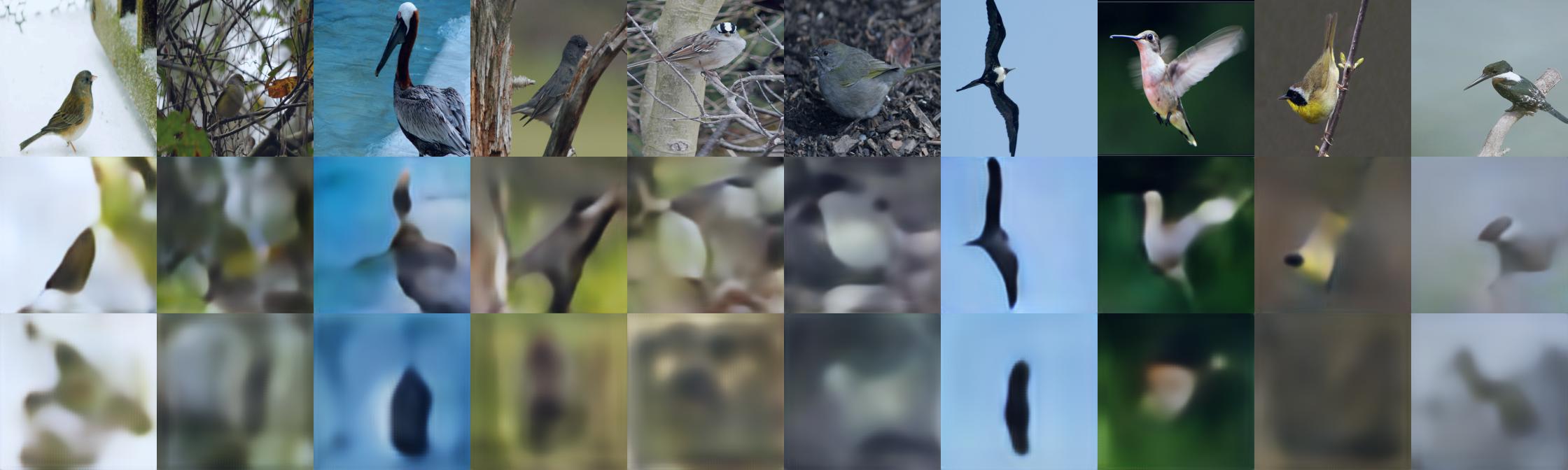}\\
\fig{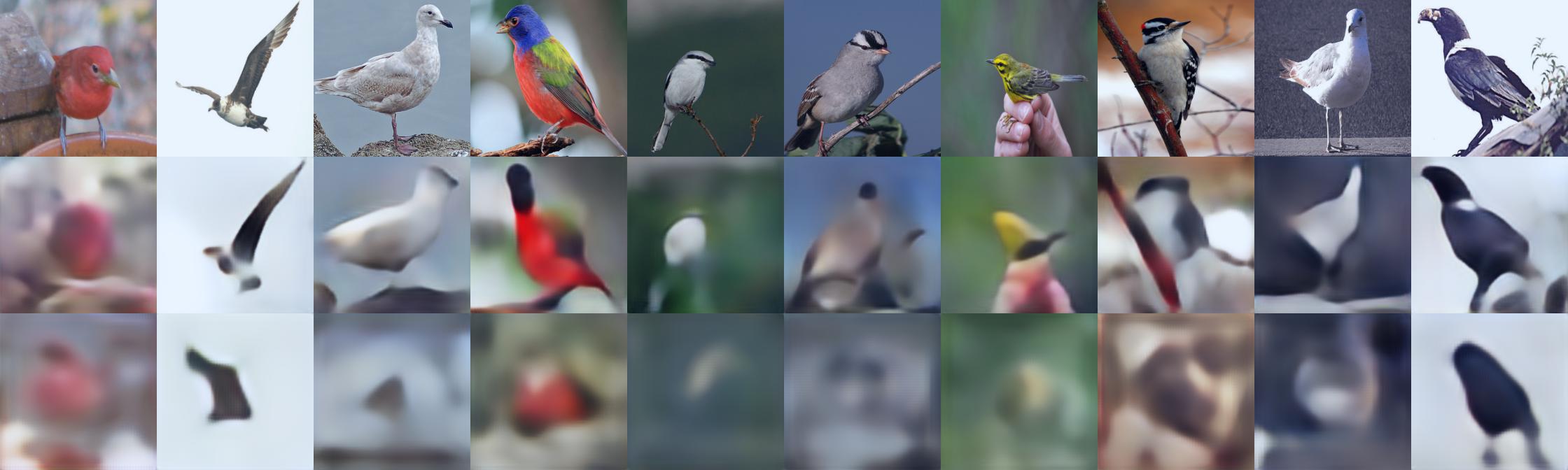}\\
\fig{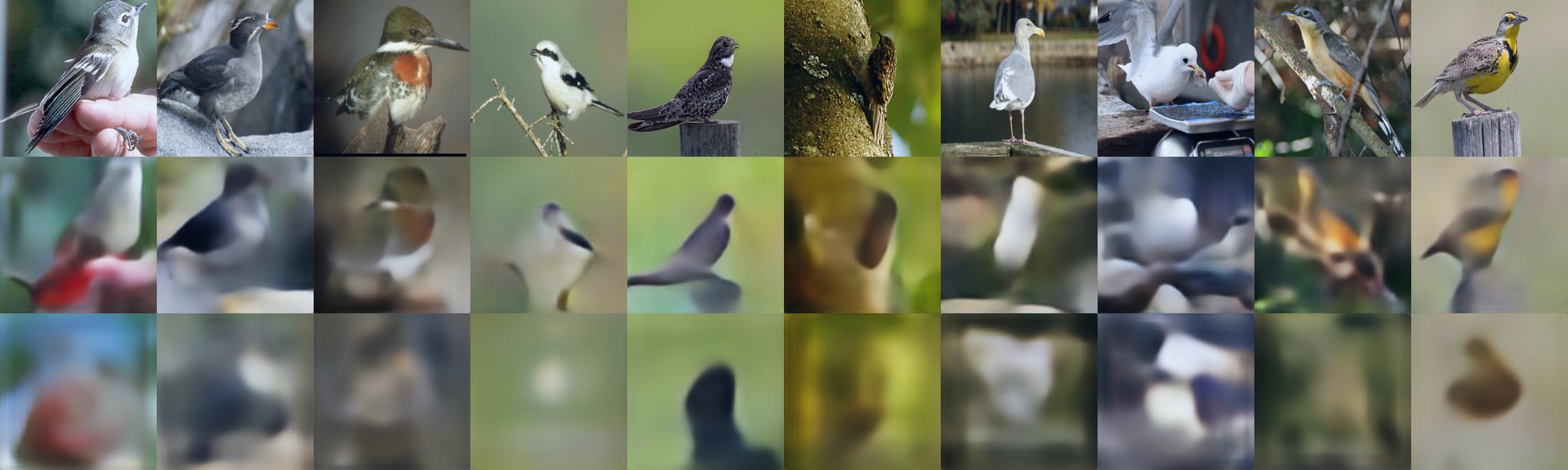}\\
\end{tabular}
\vspace{3pt}
\caption{\emph{CUB images reconstructed from tensor/vector features of original images}. Each set of 3 rows depicts the original images (row 1), followed by the images reconstructed by the tensor (row 2) and the vector (row 3) reconstructor. Meant for visualization only.}
\label{fig:mixed_big}
\end{figure*}

\begin{figure*}[ht]
\small
\centering
\setlength\tabcolsep{1pt}
\begin{tabular}{c}
\fig{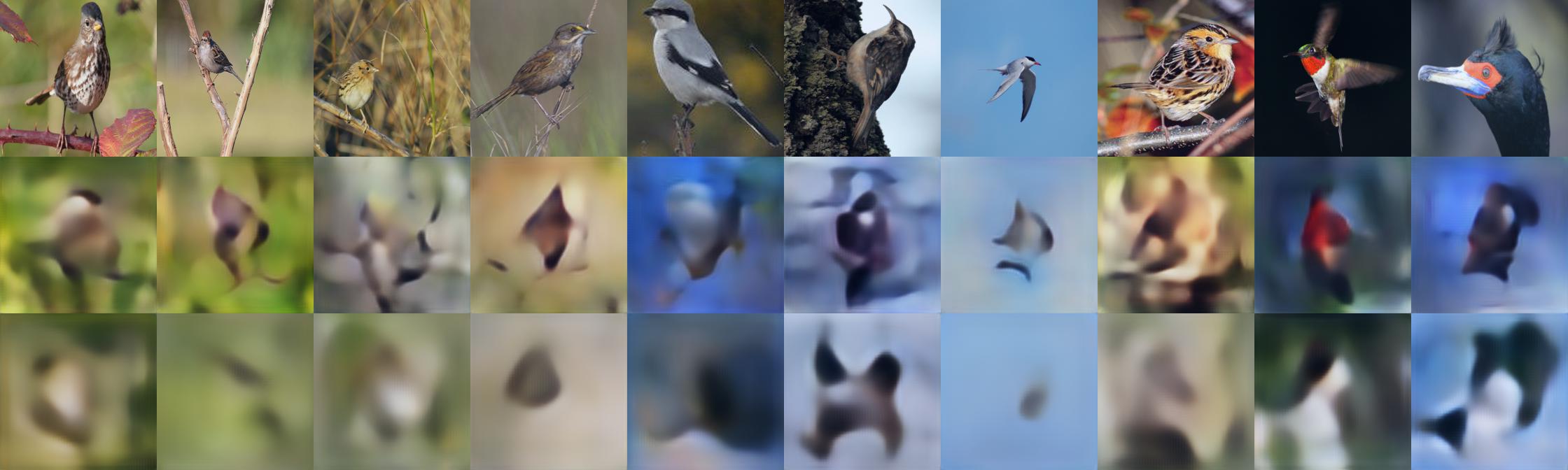}\\
\fig{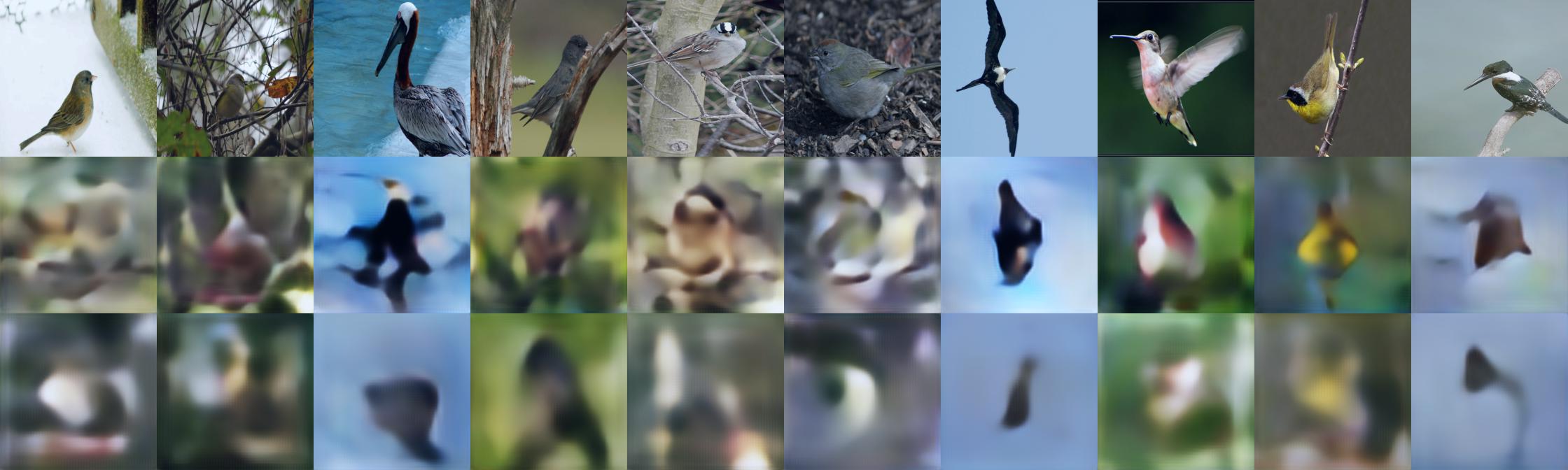}\\
\fig{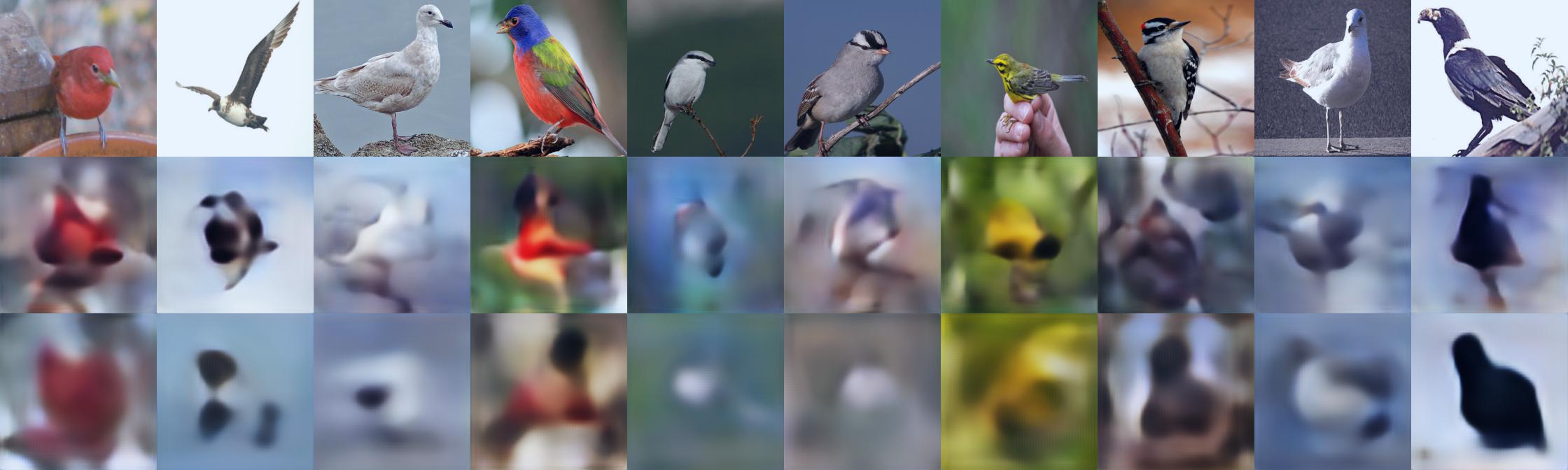}\\
\fig{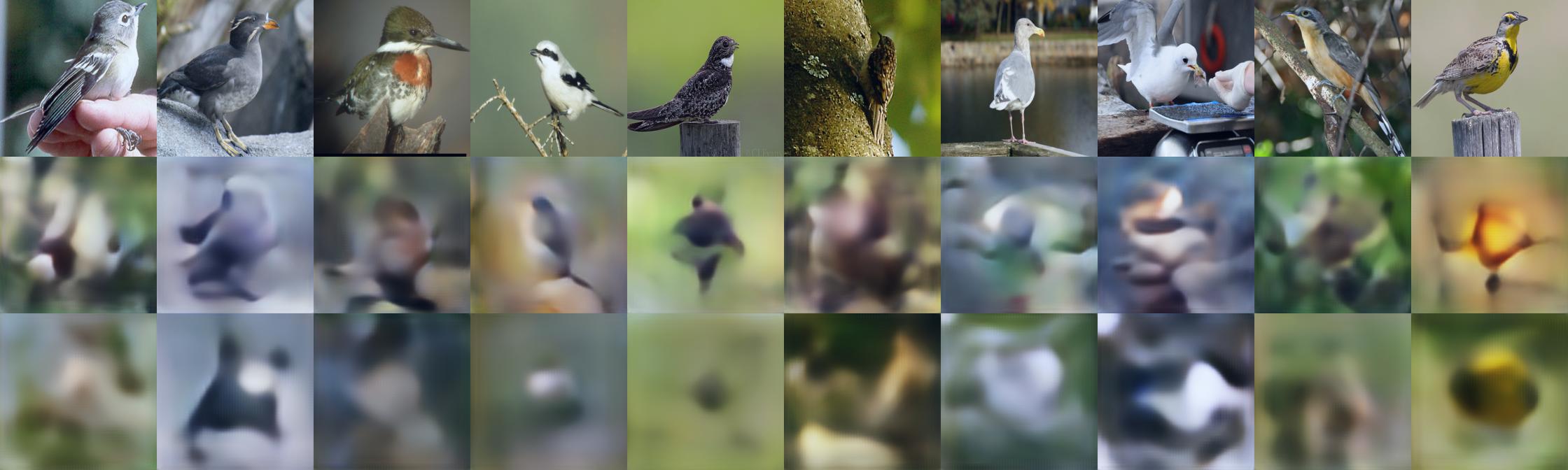}\\

\end{tabular}
\vspace{3pt}
\caption{\emph{CUB images reconstructed from our generated tensor/vector features}. Each set of 3 rows depicts the original images (row 1), followed by the images reconstructed by the tensor (row 2) and the vector (row 3) reconstructor. Meant for visualization only.}
\label{fig:generated_big}
\end{figure*}

We train each image reconstructor separately, taking as input the features as provided from the pre-trained ResNet-18 backbone, with and without GAP. For fair comparison, both reconstructors use exactly the same training settings, with $\ell_1$ reconstruction loss as the loss function, batch size 128, Adam optimizer with an initial learning rate of 0.01 and 500 epochs with learning rate decreasing by a factor of 4 every 100 epochs. Similarly to \autoref{fig:im_recon}, is evident from \autoref{fig:mixed_big} that images reconstructed from tensor features are perceptually more similar to the original. The same holds for \emph{generated} tensor and vector features, as shown in \autoref{fig:generated_big}. This experiment is for visualization purposes only; these images are not used in any way by our method.

\end{document}